\documentclass{article} 
\usepackage{iclr2022_conference,times}

\usepackage{hyperref}
\usepackage{url}

\usepackage{amsmath}
\usepackage{amsfonts}
\usepackage{amssymb}
\usepackage{wrapfig}
\usepackage{subcaption}
\usepackage{multirow}
 \usepackage{mathtools} 

\usepackage{verbatim}

\usepackage{anyfontsize}

\usepackage{microtype}
\usepackage{graphicx}
\usepackage{booktabs} 
\usepackage{multirow}
\usepackage{amsmath,amssymb}
\usepackage{booktabs}
\usepackage{caption,subcaption}
\usepackage{xcolor}
\usepackage{enumitem}
\setitemize{itemsep=10pt,topsep=0pt,parsep=0pt,partopsep=0pt}
\usepackage{adjustbox}

\newcommand{\RN}[1]{%
	\textup{\lowercase\expandafter{\it \romannumeral#1}}%
}
\usepackage{tabu}

\newcommand{\ie}[0]{\emph{i.e., }}

\newcommand{\eg}[0]{\emph{e.g., }}

\newcommand{\beq}{\vspace{0mm}\begin{equation}}
\newcommand{\eeq}{\vspace{0mm}\end{equation}}
\newcommand{\beqs}{\vspace{0mm}\begin{eqnarray}}
\newcommand{\eeqs}{\vspace{0mm}\end{eqnarray}}
\newcommand{\barr}{\begin{array}}
\newcommand{\earr}{\end{array}}

\newcommand{\xv}{\boldsymbol{x}}

\newcommand{\zv}{\boldsymbol{z}}

\newcommand{\thetav}{\boldsymbol{\theta}}

\newcommand{\R}{\mathbb{R}}

\newcommand{\Mcal}{\mathcal{M}}
\newcommand{\Lcal}{\mathcal{L}}
\newcommand{\Ocal}{\mathcal{O}}
\newcommand{\Pcal}{\mathcal{P}}

\newcommand{\Vcal}{\mathcal{V}}

\usepackage{color, colortbl}
\definecolor{Gray}{gray}{0.93}

\usepackage{xcolor,amsmath}
\usepackage[linesnumbered,ruled,vlined]{algorithm2e}
\DontPrintSemicolon

\SetKwComment{Comment}{\color{green!50!black}\# }{}

\newcommand{\var}{\texttt}
\newcommand{\FuncCall}[2]{\texttt{\bfseries #1(#2)}}
\SetKwProg{Function}{def}{:}{}

\SetKwProg{For}{for}{:}{}
\SetKwProg{If}{if}{:}{}

\newcommand{\shortname}{EsViT}

\usepackage{xcolor}
\usepackage{color, colortbl}

\definecolor{emerald}{rgb}{0.31, 0.78, 0.37}

\usepackage{tikz}
\usepackage {hyperref}

\title{Efficient  Self-supervised Vision \\Transformers for Representation Learning}

\author{Chunyuan Li$^{1}$ \,\, Jianwei Yang$^{1}$ \,\,  Pengchuan Zhang$^{1}$ \,\, Mei Gao$^{2}$ \,\, Bin Xiao$^{2}$ \,\, Xiyang Dai$^{2}$  \\ 
  \textbf{Lu Yuan}$^{2}$ \,\, \textbf{Jianfeng Gao}$^{1}$ \\
  $^1$Microsoft Research at Redmond, $^2$Microsoft Cloud + AI\\
\texttt{\{chunyl,jianwyan,penzhan,xuga,bixi,xidai,luyuan,jfgao\}@microsoft.com}
}

\iclrfinalcopy 
\begin{document}

\maketitle

\maketitle

\begin{abstract}
  
This paper investigates two techniques for developing efficient self-supervised vision transformers (\shortname{}) for visual representation learning. First, we show through a comprehensive empirical study that multi-stage architectures with sparse self-attentions can significantly reduce modeling complexity but with a cost of losing the ability to capture fine-grained correspondences between image regions. Second, we propose a new pre-training task of region matching which allows the model to capture fine-grained region dependencies and as a result significantly improves the quality of the learned vision representations. 
Our results show that combining the two techniques, \shortname{} achieves 81.3\% top-1 accuracy on the ImageNet linear probe evaluation, outperforming prior arts with around an order magnitude of higher throughput.  When transferring to downstream linear classification tasks, \shortname{} outperforms its supervised counterpart on 17 out of 18 datasets.
The code and pre-trained models are released at: \url{ https://github.com/microsoft/esvit}  


\end{abstract}


\section{Introduction}
\vspace{-2mm}
Self-supervised learning (SSL) with Transformers~\citep{vaswani2017attention} has become a de facto standard of model choice in natural language processing (NLP). The dominant approaches such as GPT~\citep{radford2018improving} and BERT~\citep{devlin2019bert} are pre-training on a large text corpus and then fine-tuning to various smaller task-specific datasets, showing superior performance. Larger Transformers pre-trained with larger-scale language datasets often lead to a stronger generalization ability, demonstrated by improved performance in downsteam tasks (with no sign of performance saturation yet), as exemplified in GPT-3~\citep{brown2020language}. 

In computer vision (CV), however, self-supervised visual representation learning is still dominated by convolutional neural networks (CNNs). Sharing a similar goal/spirit with NLP, SSL in CV aims to learn general-purpose image features from raw pixels without relying on manual supervisions, and the learned networks are expected to serve as the backbone of various downstream tasks such as classification, detection and segmentation. Recently, impressive performance have been achieved by CNN-based SSL, outperforming state-of-the-art (SoTA) fully-supervised pre-training methods~\citep{he2020momentum,caron2020unsupervised} on tasks with a limited number of labels. The key to success is view-level learning: maximizing agreement of learned representations between differently augmented views of the same example.  Recent works, including SimCLR-v2~\citep{chen2020big}, BYOL~\citep{grill2020bootstrap} and SwAV~\citep{caron2020unsupervised}, have scaled up the CNN-based models to hundreds of millions of parameters.
However, SSL has not enjoyed the same scaling success in CV as that in NLP.



Several attempts have been made to close the gap by combining SSL with Transformer and self-attention architectures. Early works include Selfie~\citep{trinh2019selfie}, which generalizes the concept of masked language modeling of BERT for images. The idea has been recently revisited in Vision Transformer (ViT)~\citep{dosovitskiy2020image} via pre-training on a much larger scale dataset, \eg JFT-300M.  ImageGPT (iGPT)~\citep{chen2020generative} generalizes the concept of auto-regressive language modeling of GPT for images, showing encouraging ImageNet recognition accuracy with a large model size. Contrastive learning with ViT has also been studied very recently in DINO~\citep{caron2021emerging} and MoCo-v3~\citep{chen2021empirical}, where new SoTA result by  linear probe evaluation on ImageNet-1K is achieved, by exhaustively consuming computation resource on full self-attention operators with long sequences of split image patches.


\begin{figure*}[t!]
	\vspace{-0mm}\centering
	\begin{tabular}{c c}
		\hspace{-4mm}
		\includegraphics[height=4.55cm]{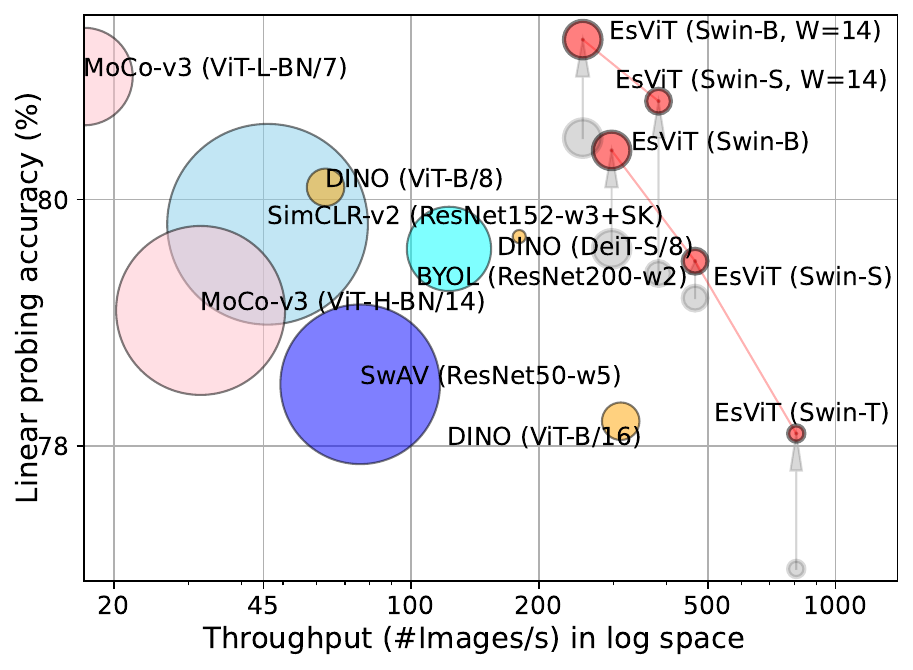}  & 
		\hspace{-0mm}
		\includegraphics[height=4.6cm]{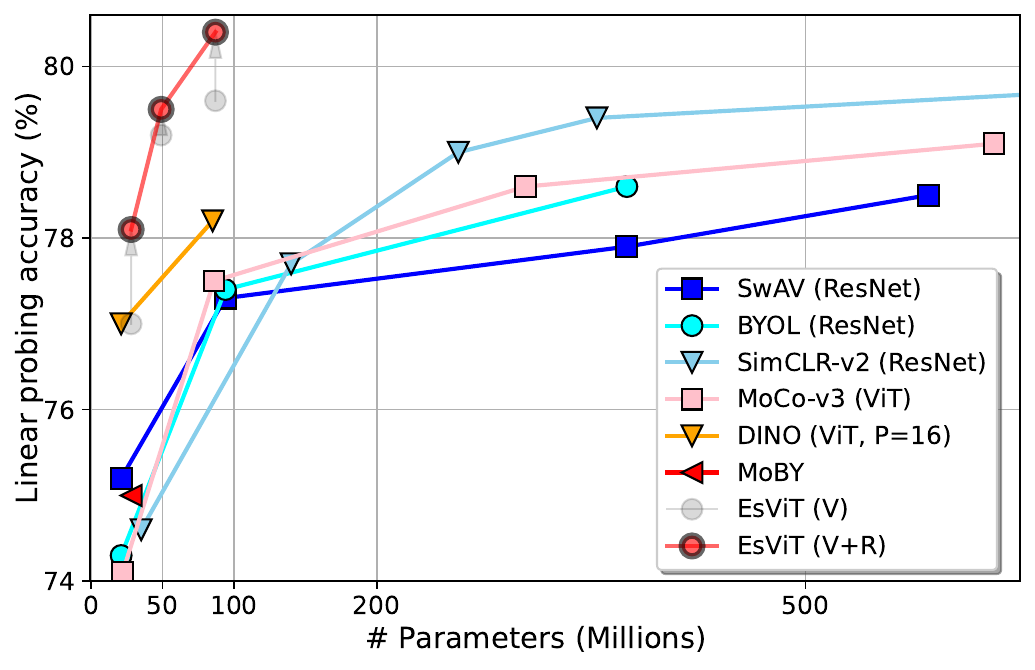} \\
	\end{tabular}
	\vspace{-3mm}
	\caption{Efficiency vs accuracy comparison under the linear classification protocol on ImageNet. Left: Throughput of all SoTA SSL vision systems, circle sizes indicates model parameter counts; Right: performance over varied parameter counts for models with moderate (throughout/\#parameters) ratio. \shortname{} pre-trained with and without the region-matching task are shown before and after the arrows, respectively. Please refer Section~\ref{sec:imagenet_probe} for details.
	 }
	\vspace{-5mm}
	\label{fig:hero_cmp}
\end{figure*}


Aiming to improve the {\em efficiency} of Transformer-based SSL, this paper presents {\em {\bf E}fficient {\bf s}elf-superivsed {\bf Vi}sion {\bf T}ransformers} (\shortname{}), by using a multi-stage architecture and a region-based pre-training task for self-supervised representation learning. 
Our main findings and contributions can be summarized as follows:


\begin{minipage}{1.0\textwidth}
\centering

\begin{enumerate}[label=(\arabic*),leftmargin=5.5mm]
\item  An intriguing property of self-supervised monolithic Transformers is firstly reported in our paper: automatic discovery of semantic correspondence between local regions. 

\item  We present the first comprehensive empirical study to show the pros and cons of multi-stage vision Transformer architectures for SSL. Though greatly reducing compute complexity, we find that the multi-stage architecture causes the loss of the property in (1).  

\item 
A region matching pre-train task is proposed to alleviate the issue in (2), and further improve the learned representations and attentions.

\item  
We validate the new \shortname{}, which combines the two techniques, on a range of tasks.
It significantly reduces the cost in building SoTA SSL vision systems, as summarized in Figure~\ref{fig:hero_cmp}, and shows better scaling performance on accuracy vs. throughput and model size.
Under the linear evaluation protocol, \shortname{} achieves 81.3\% top-1 accuracy, showing the best performance compared with all systems, and is $3.5\times$ parameter-efficient and has at least $10\times$ higher throughput than previous SoTA (81.0\%, MoCo-v3 with ViT-BN-L/7~\citep{chen2021empirical}). 
Compared with its supervised counterpart Swin Transformers~\citep{liu2021Swin}, \shortname{} shows superior performance on 17 out 18 datasets, when transferring the learned representations to downstream linear classification tasks. 

\end{enumerate}

\end{minipage}

%



\vspace{-1mm}
\section{Methods}
\vspace{-3mm}

Transformer-based SSL methods emerge very recently to lead the state-of-the-art performance on the ImageNet linear probe task~\citep{chen2021empirical,caron2021emerging}. It inherits the successes from (1) monolithic Transformer architectures that dominate in NLP~\citep{devlin2019bert,radford2018improving}, and (2) instance-level contrastive learning objectives that demonstrate arguably the best SSL performance in computer vision~\citep{chen2020simple}.
Though simple and effective, the existing Transformer-based SSL methods require a large amount of compute resources (\eg $>$1.7 TPU years of training) to reach SoTA performance.
We believe that the SSL system {\em efficiency} is highly related to two ingredients: the network architecture and the pre-train task. To strike for a better tradeoff between accuracy and efficiency, we present \shortname{}, showing better synergy of networks (a multi-stage Transformer architecture) and pre-train tasks (a non-contrastive region-matching task).



\subsection{Network Architectures: From Monolithic to Multi-stage ViT Backbone}
\vspace{-2mm}

\paragraph{Multi-stage ViT.} This paper presents the first empirical study of multi-stage Transformer architectures~\citep{vaswani2021scaling,wang2021pyramid,liu2021Swin,zhang2021vil,wu2021cvt} for SSL. Each stage consists of a {\em patch merging/embedding} module, and a {\em Transformer with sparse self-attention} module.
$(\RN{1})$ The patch merging module plays a slightly different roles in different stages. In the first stage, it splits an input RGB image into non-overlapping
patches. Each patch is treated as a ``token'', constructed as a concatenation of the raw pixel RGB values, which is further projected into a $C$-dimension feature.
In the later stage, the patch merging module concatenates the features of each group of $2\times2$ neighboring patches, and applies a linear layer on the 4$C$-dimensional concatenated features. This reduces the number of tokens by a multiple of $2\times2=4$, and the output dimension is set to $2C$.
$(\RN{2})$ A Transformer with sparse self-attention module are then employed to enable interactions among the merged features. 
The two modules above are repeated for multiple times, typically 4 times, resulting in a multi-stage ViT. As a result, a hierarchical representation is generated: the number of tokens is reduced and the feature dimension (and the number of heads in self-attentions) of each token is increased, as the network
gets deeper. 
An overview comparison of the monolithic and multi-stage Transformer architectures for SSL is illustrated in Figure~\ref{fig:illustration_esvit} in Appendix. 

\paragraph{An intriguing property of self-supervised monolithic ViT.} 
Though straightforward in implementation, changing from monolithic to multi-stage architecture without careful treatments may lose some desirable properties of self-supervised Transformers
In out study, we first empirically note an intriguing property of self-supervised monolithic ViT\citep{caron2021emerging}: {\em the pre-trained model exhibits a very strong ability to automatically discovers correspondences, even without a region-level matching objective specified in training}. 

We quantitatively evaluate the correspondence learning to illustrate this property, as discussed in the following process.
$(i)$ {\em Simulated benchmark}. Based on 50K images in the ImageNet validation dataset, we create a simple evaluation benchmark with mild augmentations: For a center-crop image, we apply \texttt{HorizontalFlip}, then \texttt{ColorJitter} and \texttt{RandomGrayscale} to create a new augmented view. In this way, ground-truth correspondences are created.
$(ii)$ {\em Evaluation process}. Given two views of the same image, we use the pre-trained backbone to extract the top-layer features, and find the feature vector in one view that best matches the other in terms of highest cosine similarity. The accuracy is measured as the averaged percentage of correctly identifying the region-to-region correspondences. Please see details in Section \ref{sec:correspondence_imagenet} in Appendix.
$(iii)$ {\em Results}.
We quantitatively show that a self-supervised monolithic ViT yields {\bf 95\%} accuracy. However, simply replacing the network with a multi-stage Transformer yields only {\bf 66\%} accuracy. This significant degradation (absolute {\bf 29\%} accuracy drop) reveals the loss of the correspondence learning property. We first raise this critical problem, and believe that it has a large impact on the pre-trained model's performance in various downstream tasks.


\vspace{-2mm}
\subsection{Pre-training Tasks: Delving into Views with Regions}
\vspace{-2mm}
We employ a non-contrastive learning framework to build our SSL method. Specifically, {\em Self-distillation with no labels} (DINO)~\citep{caron2021emerging} is considered. It leverages the knowledge distillation learning paradigm where a student network $g_{\thetav_s}$ is trained
to match the output of a given teacher network $g_{\thetav_t}$
, parameterized by $\thetav_s$ and $\thetav_t$ respectively. 
The neural network $g$ is composed of a backbone $f$ (\eg Transformers or ConvNets), and of a projection head $h$: $g = h \circ f$. The features used in downstream tasks are the output of backbone $f$.
In SSL, different augmented views $\Tilde{\xv}$ of an image $\xv$ are fed into backbone network to obtain feature maps $\zv =f(\Tilde{\xv})$. Two MLP heads followed by \texttt{softmax} per network further convert the feature vectors $z \in \zv$ into probability vectors $p=h(z)$; one head for view-level and the other head for region-level, respectively.  

More precisely, from a given image, we generate a set $\Vcal$ of different views\footnote{This set often contains views of two different resolutions $\Vcal = [\Vcal_g, \Vcal_l  ]$, where $\Vcal_g = \{ \Tilde{\xv}_{g_i} | i=1,2 \}$ is a global-view set of higher resolution, and $\Vcal_l = \{ \Tilde{\xv}_{l_i} | i=1,\dots,8 \}$  is a local-view set of lower resolution. All views $\Vcal$  are passed through the student while only the global views $\Vcal_g$ are passed through the teacher.} following~\citep{caron2021emerging}. 
The resulting feature map at the top layer for each view is $\zv = [z_1, \dots, z_T] $, where $T$ is the sequence length, and $z_i$ is a region-level representation for the local patch at position $i$. Average pooling is applied to obtain the view-level representation $\bar{z}= \text{avg-pool} (\zv)$.

\paragraph{View-level task}
Given the augmented view set for student $ \Vcal$ and teacher $ \Vcal^*$, a set of pairs $ \Pcal = \{ (s,t) | \Tilde{\xv}_s \in \Vcal,  \Tilde{\xv}_t \in \Vcal^*  \text{~and~} s \neq t $ \} is constructed to perform cross-view prediction tasks. We consider the pre-training task at the view level proposed by~\citep{caron2021emerging}:
\begin{align}\label{eq:obj_view}
\Lcal_V = \frac{1}{| \Pcal|} \sum_{ (s, t) \in \Pcal }  
\Mcal_V ( s, t), ~~\text{with} ~~
 \Mcal_V (s, t ) = -  p_{s} \log p_t,
\end{align}
where 
$p_s = h(\bar{z}_s)$ and $p_t = h(\bar{z}_t)$ are the probability output of an MLP head $h$ over the view-level representations $\bar{z}_s$ and $\bar{z}_t$, learned by student and teacher, respectively. In DINO, ViT/DeiT are considered, hence the view-level representation is the feature of the $\texttt{[CLS]}$ token.

\paragraph{Region-level task}
In~\citep{caron2021emerging}, the $\Lcal_{V}$ encourages ``local-to-global'' correspondences only at a coarse level: the large crop and the small crop are matched in the view level, leaving region-to-region correspondence unspecified. In monolithic Transformers, the drop paths and skip connections from low-level features to high-level features help the the latter to remain discriminative, thus maintain good region-matching performance. However, such a property gets diluted due to the merging operators in multi-stage Transformers. As shown in our experiments later, training a multi-stage network with $\Lcal_{V}$ only indeed results in sub-optimal representations. 

Further, it could be a waste of computation not to leverage region-level features $\zv$ that are computed in the process of extracting view-level feature.
Inspired by the success of masked language modeling task in BERT, we argue that it is important to have region-level pre-training task for computer vision, so that the model can (1) amortize the computation and fully leverage the extracted region-level features, and (2) take into account the co-occurrences/structures between local features. Unfortunately, directly performing masked patch
prediction (MPP) for the multi-stage Transformer architecture is infeasible, as the one-to-one correspondences between the input visual tokens and output features get diluted due to the merging operation. Even for monolithic architectures, MPP has not been proved effective in computer vision, as empirically shown in~\citep{dosovitskiy2020image}.

\label{sec:correspondence_imagenet}

To address this problem, we propose a non-contrastive, region-matching method
that directly works at the level of local features
by taking into account their correspondences:
\begin{align}\label{eq:obj_region}
\vspace{-2mm}
\hspace{-2mm}
 \Lcal_R = \frac{1}{| \Pcal|}  \sum_{ (s, t) \in \Pcal }  
\Mcal_R ( s, t), ~~\text{with} ~~
 \Mcal_R (s, t ) = - \frac{1}{T}\sum_{i=1}^T  p_{j^*} \log p_i, ~~ j^* = \arg \max_{j} \frac{z_i^T z_j}{\|z_i\| \|z_j\|},
\end{align}
where 
$p_i = h^{\prime}(z_i)$ and $p_j = h^{\prime}(z_j)$ are the probability outputs of a new MLP head $h^{\prime}$ over the local features of student $z_i \in \zv_s$ and teacher $z_j \in \zv_t$, respectively. $j^*$ is the index of the feature in $\zv_t$ that best matches the $i$-th feature in $\zv_s$, in the sense of highest cosine similarity. Note that $z_i$ and $z_{j^*}$ are {\em contextualized} features of two best matched regions from different augmentated views, minimizing $\Lcal_R$ encourages different contexts (\ie surrounding regions) to learn invariant features, and thus captures the region-dependency.

\begin{wrapfigure}{R}{0.5\textwidth}
\vspace{-5mm}
\begin{minipage}{0.5\textwidth}
\scriptsize
\centering
\hspace{-3mm}
\includegraphics[height=3.2cm]{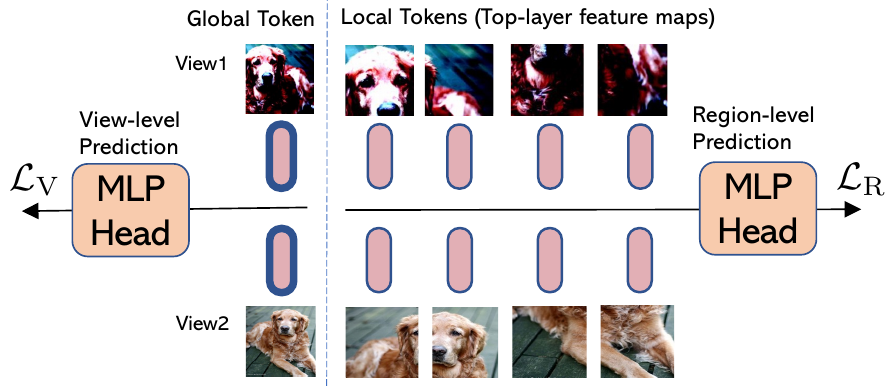}
\vspace{-1mm}
\caption{Pre-training objectives, including view-level  (left) and region-level (right) prediction.} 
\label{fig:esvit_preidctions}
\vspace{-3mm}
\end{minipage}
\end{wrapfigure}
The overall pre-training objective of \shortname{} is $\Lcal = \Lcal_R + \Lcal_V$, we learn to match the feature distributions at both the view and region levels by minimizing the cross-entropy loss w.r.t. the parameters of the student network $g_{\thetav_s}$. A visual illustration is in Figure~\ref{fig:esvit_preidctions}, and the full algorithm is in Appendix. 
We updates teacher/student network alternatively:
$(\RN{1})$ Given a fixed teacher network, the student network is updated by minimizing the full cross-entropy loss:
$\thetav_s  \leftarrow \arg \min_{\thetav_s} \Lcal(s, t ; \thetav_s)$.
$(\RN{2})$ The teacher model is updated as an exponential moving average (EMA) of the student weights
$\thetav_t  \leftarrow \lambda \thetav_t + (1 - \lambda) \thetav_s$, with $\lambda$  following a cosine schedule
from 0.996 to 1 during training.
By default, the full objective $\Lcal$ is used from the beginning. One can also load a checkpoint trained by $\Lcal_V$ only, and add $\Lcal_R$ for continual pre-training, which is shown effective in boosting performance in our experiments.

\paragraph{Computational overhead}
Note that applying $\Lcal_R$ on the traditional monolithic Transformer architecture can be prohibitively computationally expensive, as it requires $\Ocal(T^2)$ to compute $\Lcal_R$. For a typical image of resolution $224 \!\times \!224$, the feature map length  of ViT/DeiT (with patch size 16) at the top layer is $T=196$, while the multi-stage architecture yields $T=49$, which requires 3 times less compute in computing $\Lcal_R$. To empirically illustrate this, we show in Appendix Section~\ref{sec:lr_efficiency} that $\Lcal_R$ adds acceptable extra memory and computational cost (around 1.2 and 1.05 $\times$, respectively) for multi-stage Transformers, while it will quickly go out-of-memory for monolithic Transformers when the batch size is increased.

\vspace{-3mm}
\section{Related Works}
\vspace{-3mm}

\paragraph{Relation to mask prediction tasks} We can consider the proposed $\Lcal_R$ as a proxy to mimick masked language modeling in BERT, where the ``ground-truth'' local token is a soft label provided by the teacher network, while the student network makes predictions to match that target, based on the context of regions in a different augmented view. Importantly, our $\Lcal_R$ considers \texttt{softmax} with cross-entropy in the objective, rather than \texttt{MSE} as in MPP. A very sharp teacher distribution is used by choosing small temperatures. This encourages the model to focus on the salient dimensions, rather than waste modeling capability on training short-range dependencies and high-frequency details~\citep{ramesh2021zero}.

\paragraph{Relation to DenseCL} The proposed $\Lcal_R$ mostly related to DenseCL~\citep{wang2020dense} in that the region correspondences in both methods are determined as the two most similar grid features. One critical difference is that DenseCL is a contrastive region-matching task, while our $\Lcal_R$ is a non-contrastive region-matching task, where no negative samples/queue is needed. This technical difference has a significant impact on the downstream task performance. We find that $\Lcal_R$ is particularly effective in serving our goal to improve image classification performance and build efficient \& affordable SoTA SSL system; In contrast, DenseCL degrades the classification performance. 

\paragraph{Relation to other region-level tasks} The ideas of leveraging local region-level pre-training tasks for visual representation learning have been explored for ConvNets~\citep{misra2020self,xiong2020loco,wang2020dense,xie2021detco,yang2021instance,xie2021propagate}. We summarize the differences in three aspects:
$(\RN{1})$
Motivation. Our region-matching task $\Lcal_R$ aims to recover the lost property of automatic correspondence learning in self-supervised monolithic Transformers, while most existing region-level tasks aim to improve dense visual prediction tasks.
$(\RN{2})$ 
Technical difference. Our $\Lcal_R$ is a non-contrastive region-matching task, while others are contrastive learning.
$(\RN{3})$
Empirical performance. Most region-level tasks improve dense visual prediction tasks but sacrifice their image classification performance, while $\Lcal_R$ consistently improves classification performance. Among them, \shortname{} training method achieves the best ImageNet linear probe performance with minimum computational overhead. For detailed comparisons, please refer to Table~\ref{table:region_level_tasks} in Appendix.

\paragraph{Self-supervised vision Transformers.} The research on Transformer-based self-supervised representation learning just scratches the tip of the iceberg, and only a few attempts are made on this topic.  ImageGPT~\citep{chen2020generative} and MoCo-v3~\citep{chen2021empirical} dedicate huge compute resource with large models to exploring the frontier. 
DINO~\citep{caron2021emerging} achieves comparable performance of large self-supervised ConvNets using small/medium-size Transformers. The proposed \shortname{} further pursues efficient and affordable solutions to self-supervised vision Transformers.
For more general related works on Transformers for vision tasks and self-supervised ConvNets, please refer to Section~\ref{sec:related_work_appendix} in Appendix. 



\vspace{-3mm}
\section{Experimental Results}
\vspace{-3mm}
We describe the experimental settings in Appendix Section~\ref{sec:experimental_settings}, and evaluate the proposed \shortname{}{} to answer three questions:
\textbf{\texttt{Q1}}: How does \shortname{} perform on standard ImageNet benchmark compared to SoTA methods?
\textbf{\texttt{Q2}}: How effective \shortname{} is when transferring to downstream tasks? 
\textbf{\texttt{Q3}}: What are the design choices and empirical contributions of $\Lcal_R$? 
\textbf{\texttt{Q4}}: When does the intriguing property of self-supervised Transformers exist, including learned correspondence and attentions?

\begin{table}[t!]
\footnotesize 
\centering
\scalebox{0.89}{
\begin{tabular}{ @{\hspace{-0pt}}l@{\hspace{8pt}}c@{\hspace{2pt}}@{\hspace{8pt}}c@{\hspace{8pt}}c@{\hspace{12pt}}l@{\hspace{12pt}}l}
\toprule
 Method   &  \#Parameters $\downarrow$~~ &  Throughput $\uparrow$ & Linear $\uparrow$~~ & $k$-NN $\uparrow$  \\ 
\hline
\multicolumn{3}{l}{\em \hspace{-3mm} SoTA SSL methods with Big ConvNets}  & & \\
SwAV, RN50w5~\citep{caron2020unsupervised} & 586 & 76 & 78.5 &  67.1 \\
BYOL, RN200w2~\citep{grill2020bootstrap} & 250 & 123 & 79.6 & 73.9 \\
SimCLR-v2, RN152w3+SK ~\citep{chen2020big} & 794 & 46 & 79.8 & 73.1 \\
\midrule
\multicolumn{5}{l}{\em \hspace{-3mm} Skyline methods with excessively long sequences for self-attentions}  \\
DINO, DeiT-S/8~\citep{caron2021emerging} & 21 & 180 & 79.7 & 78.3 \\
DINO, ViT-B/8~\citep{caron2021emerging} & 85 & 63 & 80.1 & 77.4 \\
MoCo-v3, ViT-B-BN/7~\citep{chen2021empirical} & 85 & $\sim$63 & 79.5 & - \\
MoCo-v3, ViT-L-BN/7~\citep{chen2021empirical} & 304 & $\sim$17 & 81.0 & - \\
iGPT, iGPT-XL~\citep{chen2020generative} &  6801 & - & 72.0 & - \\
\rowcolor{Gray}
\shortname{}, Swin-T/$W\!=\!14$ & 28 & 660   &  78.7 \textcolor{blue}{\scriptsize (77.9)}  &   77.0  \textcolor{blue}{\scriptsize (75.5)} \\
\rowcolor{Gray}
\shortname{}, Swin-S/$W\!=\!14$ & 49 & 383   &  80.8 \textcolor{blue}{\scriptsize (79.4)}  &   79.1 \textcolor{blue}{\scriptsize (77.3)} \\
\rowcolor{Gray}
\shortname{}, Swin-B/$W\!=\!14$ & 87 & 254   &  {\bf 81.3} \textcolor{blue}{\scriptsize (80.5)}  &  {\bf 79.3}  \textcolor{blue}{\scriptsize (78.3)} \\
\midrule
\multicolumn{5}{l}{\em \hspace{-3mm} Transformer-based SSL, with moderate sequence length for self-attentions}  \\
\textcolor{gray}{Masked Patch Pred., ViT-B/16~\citep{dosovitskiy2020image}} & \textcolor{gray}{85} & \textcolor{gray}{312} &  ~\textcolor{gray}{79.9}$^{\dagger}$ & - \\
DINO, DeiT-S/16~\citep{caron2021emerging} & 21 & 1007 & 77.0 & 74.5 \\
DINO, ViT-B/16~\citep{caron2021emerging} & 85 & 312 & 78.2 & 76.1 \\
MoCo-v3, ViT-B/16~\citep{chen2021empirical} &  85  & 312 & 76.7 & - \\
MoCo-v3, ViT-H-BN/16~\citep{chen2021empirical} & 632 & $\sim$32 & 79.1 & - \\
MoBY, Swin-T~\citep{xie2021moby} & 28 &  808 & 75.1 & - \\
\rowcolor{Gray}
\shortname{}, Swin-T & 28 &  808 &  78.1 \textcolor{blue}{\scriptsize (77.0)}  & 75.7 \textcolor{blue}{\scriptsize (74.2)} \\
\rowcolor{Gray}
\shortname{}, Swin-S & 49 &  467 &  79.5  \textcolor{blue}{\scriptsize (79.2)}  & 77.7 \textcolor{blue}{\scriptsize (76.8)}\\
\rowcolor{Gray}
\shortname{}, Swin-B & 87 &  297 &  {\bf 80.4} \textcolor{blue}{\scriptsize (79.6)} & {\bf 78.9} \textcolor{blue}{\scriptsize (77.7)} \\
\bottomrule
\end{tabular} }


\vspace{0mm}
\caption{Comparison with SoTA across different architectures on ImageNet linear probing. \shortname{} with $\Lcal_L + \Lcal_R$ is reported, while \shortname{} with only $\Lcal_R$ is shown in parentheses. $W=14$ is the window size, otherwise the default $W=7$. ViT-BN is ViT that has BatchNorm~\citep{frankle2020training}, and “$/P$” denotes a patch size of $P\!\times\!P$.  ``$\sim$'' indicates through-puts estimated by comparing different papers, detailed in Appendix. $^{\dagger}$ The mask patch prediction in~\citep{dosovitskiy2020image} is pre-trained on JFT-300M and end-to-end fine-tuned in ImageNet, which we append as a reference.}
\label{tab:main_result_cls}
\vspace{-5mm}
\end{table}






\vspace{-3mm}
\subsection{Comparisons with Prior Art on ImageNet}
\label{sec:imagenet_probe}
\vspace{-3mm}
We report top-1 linear probe and $k$-NN accuracy on the ImageNet validation set.
Table~\ref{tab:main_result_cls} presents comparisons with SoTA SSL systems across various architectures. Please refer to Figure~\ref{fig:hero_cmp} for comparisons over scaling parameter counts and throughput. Our findings are summarized below.

\paragraph{Comparisons with self-supervised Transformers.} 


The DINO- and MoCo-based ViT has higher accuracy and smaller models than iGPT, under the same linear probing protocol and training data. At the similar level of model size and compute complexity, the proposed \shortname{} improve SoTA methods DINO/MoCo-v3 by a large margin: \shortname{} (Swin-B) outperforms  DINO (ViT-B/16) by 2.2\% linear probe accuracy and 2.8\% $k$-NN accuracy in absolute values.
\shortname{} (Swin-B) even performs slightly better than DINO (ViT-B/8) (0.3\% higher linear probe accuracy and 1.5\% higher $k$-NN accuracy), with $4\times$ higher throughput.
MoBY~\citep{xie2021moby} is a con-current work that investigates multi-stage ViT in SSL. With the same architecture Swin-T, our \shortname{} pre-training tasks significantly outperform MoBY, showing 3\% higher accuracy. 
In \shortname{}, longer sequences in self-attention is implemented by increasing the window size. We experiment this by considering a window size of $W\!=\!14$. 
Overall, the proposed \shortname{} (Swin-B/$W$=14) shows the best performance (top-1 accuracy 81.3\%, top-5 accuracy 95.5\%, $k$-NN accuracy 79.3\%), compared with all systems, and is $3.5\times$ parameter-efficient and has at least $10\times$ higher throughput than previous SoTA MoCo-v3.



\paragraph{Comparisons with big ConvNets.}
We compare with the SoTA big ResNets reported by SimCLR-v2~\citep{chen2020big}, BYOL~\citep{grill2020bootstrap} and SwAV~\citep{caron2020unsupervised}. Among them, the best accuracy $79.8\%$ under the linear probing protocol is reported by SimCLR-v2 with SK-ResNet, where Selective Kernel (SK)~\citep{li2019selective} is a form of attention to enhance CNNs. 
It is clear in Figure~\ref{fig:hero_cmp} (b) that all ConvNets-based SSL methods show an envelope in the regime of scaling up model sizes after passing $500$M.
\shortname{} achieves better accuracy than their highest envelope, with $16\times$ less model parameters and $8\times$ higher throughput.


\vspace{-2mm}
\subsection{Transfer Learning}
\vspace{-2mm}
We also conduct transfer learning in downstream tasks to evaluate the quality of learned representations. Two sets of tasks are considered:

\begin{minipage}{0.99\textwidth}
\centering

\begin{itemize}[leftmargin=5.5mm]
\item {\em Classification on a suite of 18 small datasets}. As exemplified in \citep{radford2021learning}, it is a common and clean approach to evaluate a learned representation by fitting a linear classifier on
the representation and measuring its performance across multiple datasets.  We study 18 datasets used in~\citep{radford2021learning}. Automatic hyper-parameter tuning is considered to ensure fairness of comparison. Besides averaged $\texttt{scores}$, we report $\texttt{\#\!\! wins}$ as the number of datasets on which the model outperforms its supervised counterpart. Detailed dataset description and settings are in Appendix.
\end{itemize}

\end{minipage}

\begin{minipage}{0.99\textwidth}
\centering

\begin{itemize}[leftmargin=5.5mm]
\item {\em Detection and segmentation on COCO}. Different from previous monolithic self-supervised ViT, the multi-stage architecture in \shortname{} can be readily used for dense visual tasks that require hierarchical feature representations.
\end{itemize}
\vspace{-2mm}
\end{minipage}

\begin{figure*}[t!]
\vspace{-0mm}\centering 
\centering
\hspace{-3mm}
\includegraphics[height=3.6cm]{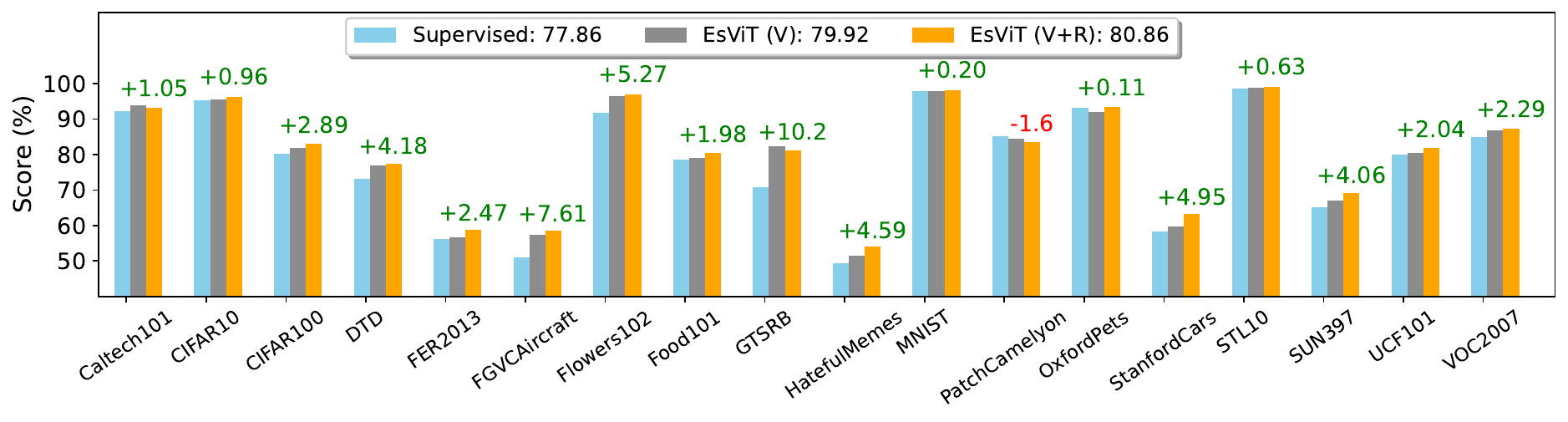}
\vspace{-2mm}
\caption{ Linear probing on 18 downstream datasets. Averaged scores are reported for each method. \shortname{} outperforms its supervised  counterpart on 17 out of 18 datasets.}
\label{fig:esvit_transfer_cls}
\vspace{-2mm}
\end{figure*}

\paragraph{Comparison with supervised counterparts.} We compare with the supervised-learning Swin, whose checkpoints are downloaded from the official codebase\footnote{\url{https://github.com/microsoft/Swin-Transformer}}. 
Figure~\ref{fig:esvit_transfer_cls} shows the classification results of Swin-S, \shortname{} consistently outperforms its supervised variant, often by a large margin. Similar conclusions are drawn for other model sizes. On COCO detection and segmentation task, however, \shortname{} shows comparable results with the variant with $\Lcal_V$ only (shown in parentheses) and the supervised counterpart (Swin-T trained with 3$\times$ schedule), as shown in Table~\ref{table:coco_res}. We hypothsize this is related to the non-constrastive nature of \shortname{}, as explained later.

\begin{table}[t!]
\begin{minipage}[b]{0.44\linewidth}

  \centering
  \scalebox{0.92}{
  \begin{tabu}{l |  lll }
    \toprule
   & AP$^{bb}$ & AP$^{bb}_{50}$ & AP$^{bb}_{75}$   \\ \cline{2-4}
    \rowfont{\color{gray}}
    Sup. &  46.0 & 68.1 & 50.3   \\
    \shortname{}  
    & 46.2  \textcolor{blue}{\scriptsize (46.2)}  
    & 68.0  \textcolor{blue}{\scriptsize (67.9)}  
    & 50.6  \textcolor{blue}{\scriptsize (50.5)}  \\
    \midrule
   &  AP$^{mb}$ & AP$^{mb}_{50}$ & AP$^{mb}_{75}$  \\ \cline{2-4}
    \rowfont{\color{gray}}
    Sup. & 41.6 & 65.1 & 44.9  \\
    \shortname{} 
    & 41.6  \textcolor{blue}{\scriptsize (41.7)}  
    & 64.9  \textcolor{blue}{\scriptsize (64.8)}  
    & 44.8  \textcolor{blue}{\scriptsize (45.1)}   \\ 
    \bottomrule
  \end{tabu}
  }
  \vspace{-2mm}
  \caption{COCO Detection \& Segmentation.}
  \vspace{-2mm}
  \label{table:coco_res}  
\end{minipage}\hfill
\begin{minipage}[b]{0.56\linewidth}
  \centering
  \scalebox{0.83}{
  \begin{tabu}{l | cc | cc}
    \toprule
    Pre-train Data    
    &   \multicolumn{2}{c|}{ImageNet-1K}  
    & \multicolumn{2}{c}{18 Datasets}    \\
          &   Linear & $k$-NN & Scores & \# Wins \\
    \midrule
    \rowfont{\color{gray}}
    Supervised & - & -  &  77.29 & - \\
    ImageNet-1K &  78.0 {\footnotesize (77.1)}  & 75.7 {\footnotesize (73.7)}  & 80.66  & 16 \\
    WebVision-v1    & 75.9 {\footnotesize (75.4)} &  71.2 {\footnotesize (69.4)} & 80.00 & 14 \\ 
    OpenImages-v4    & 70.6 {\footnotesize(69.6)} &  62.0 {\footnotesize (60.3)}   & 77.97 & 10 \\
    ImageNet-22K   & 75.0 {\footnotesize(73.5)} & 67.9 {\footnotesize (66.1)}  & {\bf 81.03} & {\bf 17} \\    
    \bottomrule
  \end{tabu}
  }
  \vspace{-2mm}
  \caption{Impact of the pre-train datasets.}
  \vspace{-2mm}
  \label{table:datasets_res}  
\end{minipage}
  \vspace{-8mm}
\end{table}

\paragraph{Effects of larger, less-curated pre-train datasets.}
The performance of Transformer-based SSL research has thus far been limited to highly curated pre-train data such as ImageNet-1K. To push the frontier in leveraging large amounts of unlabeled data, we explore the effects of pre-training from larger, less-curated image datasets:  
WebVision-v1~\citep{li2017webvision}, OpenImages-v4~\citep{kuznetsova2020open} and ImageNet-22K~\citep{deng2009imagenet}, described in Appendix. The pre-train epochs on different datasets are adjusted so that all models see a similar number of augmented views. 
We summarize the results in Table~\ref{table:datasets_res} and would like to emphasize the following findings.
First, $\Lcal_R$ improves $\Lcal_V$ (shown in parentheses) on all datasets.
Second, all \shortname{} pre-trained checkpoints outperform supervised checkpoint in downstream classification tasks, but performance varies a lot, with ImageNet-22K checkpoint showing the best transfer ability. Third, ImageNet-1K pre-trained model shows the best ImageNet-1K linear probe performance. 
We hypothesize that it is not only the size of pre-train dataset matters, but also the distribution of image classes matters: more diverse and well-balanced distribution results in a stronger generalization ability. 


\vspace{-2mm}
\subsection{Discussion on the Non-contrastive Region-Matching Task}
\vspace{-2mm}

\paragraph{Compatibility with various network architectures.} We investigate ResNet-50 and different efficient sparse Transformers in Table~\ref{tab:cmp_attn_appendix}. DeiT is shown as a baseline reference. Batch size = 1024 in this experiment. To ensure fair comparison, we modify all into a 4-stage architecture with the number of Transformer layers in each stage as 2-2-6-2. We see that $\Lcal_R$ improves all network architectures, including ResNet-50, Swin~\citep{liu2021Swin}, ViL~\citep{zhang2021vil}, CvT~\citep{wu2021cvt} and PvT~\citep{wang2021pyramid}. 
Though directly adding $\Lcal_R$ to monolithic ViT is computationally infeasible, we uniformly sampled top-layer grid features of DeiT and then add $\Lcal_R$, but did not observe performance improvement. This is partly because the monolithic ViT itself already has a good corresponding ability, an extra region-matching task does not provide new learning signals.  As compared in Appendix Table~\ref{table:r50_linear_probe} with the ResNet-50 backbone, \shortname{} learning method shows the highest accuracy, compared with existing SSL methods.


\paragraph{Model scaling with $\Lcal_R$.} We compare the pre-training objective with and without $\Lcal_R$ in Table~\ref{tab:main_result_cls}. Across different model scales and window sizes, the proposed region level $\Lcal_R$ can consistently improve the performance. The gains can be clearly seen by $k$-NN accuracy (around 1-2\%),  where no additional tuning is needed as in linear probe. Figure~\ref{fig:obj_curves} demonstrates that $\Lcal_R$ helps model convergence, and can be used as a drop-in to improve models trained with the view level task.

\paragraph{Contrastive {\em vs} Non-contrastive region-matching tasks.} The proposed $\Lcal_R$ adds a non-contrastive region-matching task to the non-contrastive view-level task $\Lcal_V$; On the contrary, DenseCL adds a contrastive region-matching task to the contrastive view-level task MoCo-v2. In Table~\ref{table:region_matching}, we compare four methods in the same setting with ResNet-50. DenseCL improves dense visual prediction performance, but hurts classification performance. $\Lcal_R$ improves both tasks, especially the classification performance. One limitation is that the non-contrastive methods show lower performance in dense prediction tasks, this is consistent with the observations for BYOL in~\citep{wang2020dense}.
The simple $\Lcal_R$ shows the best ImageNet accuracy compared with all sophisticated region-level tasks in this 200-epoch setting in Appendix Table~\ref{table:region_level_tasks}, and the best overall accuracy in Table~\ref{table:r50_linear_probe}. It indicates that $\Lcal_R$ well serves our goal in building efficient SoTA SSL systems.

\paragraph{Design choices of $\Lcal_R$.} We ablate a couple of choices in constructing $\Lcal_R$ in Eq. \eqref{eq:obj_region}. 
$(\RN{1})$ \texttt{Softmax} {\em vs} \texttt{MSE}. One alternative way to measure the distance between two projected vectors is \texttt{MSE}, as employed in the popular non-contrastive SSL algorithm BYOL~\citep{grill2020bootstrap}. When adding region-matching tasks to BYOL and pre-training 50 epochs, \texttt{Softmax} and \texttt{MSE} yield $k$-NN accuracy of 37.2\% and 34.9\%, while the baseline BYOL yields 33.1\%. We also replace the region-matching metric in \shortname{} as \texttt{MSE}, yielding $k$-NN accuracy 72.6\%, which lower than the view-level task only (74.2\%). These results show that  \texttt{Softmax} is essential in $\Lcal_R$.
$(\RN{2})$ Optimal Transport (OT) {\em vs} Simple Argmax. To avoid heavy computational overhead, a simple feature-level argmax solution is considered in Eq. \eqref{eq:obj_region} to pair two local regions. To study the impact of high region-matching quality, we consider OT. Empirically, we observe OT yields slightly higher $k$-NN accuracy at the early stage, but the gain is diminished in the end. Considering the extra computational cost of solving OT with an inner loop in sinkhorn algorithm~\citep{cuturi2013sinkhorn}, we opt for simple argmax in our experiments.   

\begin{table}[t!]


\begin{minipage}[b]{0.50\linewidth}

\scalebox{0.70}{
\begin{tabular}{ @{\hspace{8pt}}c@{\hspace{2pt}}@{\hspace{5pt}}c@{\hspace{5pt}}c@{\hspace{5pt}}|c@{\hspace{8pt}}|c@{\hspace{8pt}}c@{\hspace{8pt}}}
\toprule
\multirow{1}{*}{Method}    &  \multirow{1}{*}{\#Param.}  &   \multirow{1}{*}{ Im./s} &  Pre-train tasks   &   Linear & $k$-NN  \\
\midrule
DeiT & 21 &  1007 & $\Lcal_{V} $  & 75.9  &  73.2 \\ \midrule
\multirow{3}{*}{R-50} & \multirow{3}{*}{24} &  \multirow{3}{*}{1237}   &  $\Lcal_{V}$  &  ~75.3$^{\dagger}$ & ~67.5$^{\dagger}$ \\
 &  &   &   $\Lcal_{V}$  & 75.0 &  69.3 \\ 
 &  &   &   $\Lcal_{V}\!+\!\Lcal_{R}$    & {\bf 75.7}  &  {\bf 71.2} \\ \midrule
\multirow{2}{*}{Swin} & \multirow{2}{*}{28} &  \multirow{2}{*}{808}   &  $\Lcal_{V} $  & 77.1 &  73.7 \\
 &  &   &   $\Lcal_{V}\!+\!\Lcal_{R}$    & {\bf 77.6}  &  {\bf 75.4} \\ \midrule
\multirow{2}{*}{ViL} & \multirow{2}{*}{28} &  \multirow{2}{*}{386}     &  $\Lcal_{V} $  & 77.3 &  73.9 \\
 &  &    &   $\Lcal_{V}\!+\!\Lcal_{R}$    & {\bf 77.5}  &  {\bf 74.5}  \\ \midrule
\multirow{2}{*}{CvT} & \multirow{2}{*}{29} &  \multirow{2}{*}{848}  
 &   $\Lcal_{V} $  & 77.6  & 74.8  \\
 &  &    &  $\Lcal_{V}\!+\!\Lcal_{R}$   &  {\bf 78.5}   & {\bf 76.7}  \\ \midrule
\multirow{2}{*}{PvT} & \multirow{2}{*}{24} &  \multirow{2}{*}{851}  
 &   $\Lcal_{V} $  & 75.4  &  72.0 \\
 &  &    &  $\Lcal_{V}\!+\!\Lcal_{R}$   &  {\bf 76.3}   & {\bf 72.9}  \\

\bottomrule
\end{tabular}} \\
\vspace{-1mm}
\caption{Different architectures with and without $\Lcal_{R}$. DeiT and ResNet-50 are shown as references. ~~$^{\dagger}$ Numbers reported in~\citep{caron2021emerging}. }
\label{tab:cmp_attn_appendix}
\end{minipage}
\hfill
\begin{minipage}[b]{0.46\linewidth}
\centering
\includegraphics[height=4.7cm]{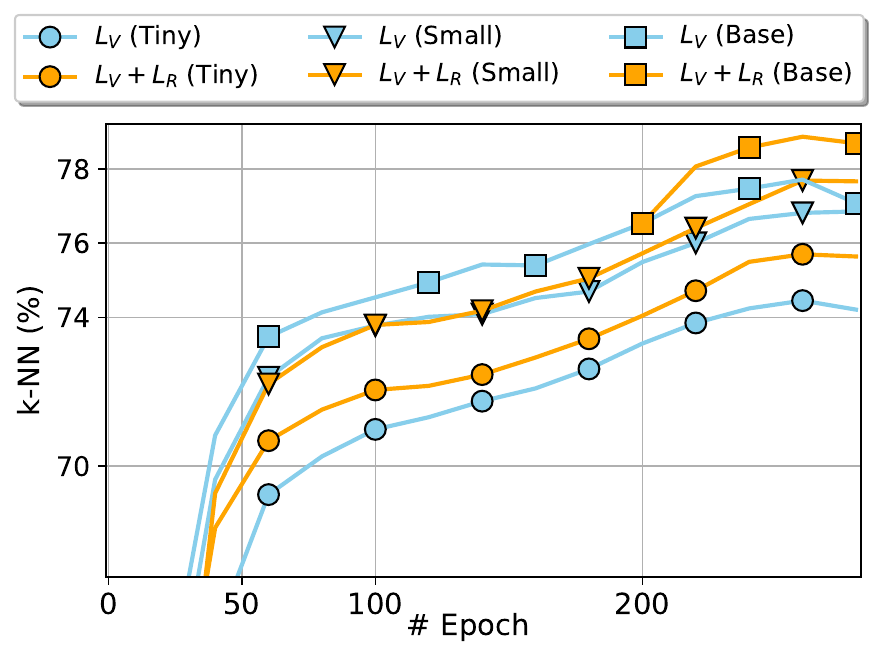}
\vspace{-4mm}
\captionof{figure}{Learning curves of different pre-training tasks. For Base model,  $\Lcal_{\text{R}}$ is added from the 200th epoch.}
\label{fig:obj_curves}
\end{minipage}
\vspace{-2mm}
\end{table}


\begin{table}[t!]
  \centering
  \centering
  \scalebox{0.83}{
  \begin{tabu}{ll cc | ll | ll}
    \toprule
 \multicolumn{4}{c|}{Pre-training {\bf ResNet50} in different settings}  
   & \multicolumn{2}{c|}{ImageNet-1K}  
   & \multicolumn{2}{c}{COCO}   \\
  Types & Methods &     \#Epochs  & \#Views  &   Linear & $k$-NN & AP$^{bb}$ &  AP$^{mb}$ \\
    \midrule
    \rowfont{\color{gray}}
   & Supervised & & & - & -  &  38.2 & 33.3 \\ \midrule
\multirow{2}{*}{Contrastive}     
   & MoCo-v2 & 200 & 2 & 67.5 & 55.6  & 38.7  & 33.9 \\
   & DenseCL & 200 & 2 & 63.6 {\bf \textcolor{red}{(-3.9)}} & 48.6 {\bf \textcolor{red}{(-7.0)}} & 39.1 {\bf \textcolor{emerald!80}{(+0.4)}}  & 34.2 {\bf \textcolor{emerald!80}{(+0.3)} }\\ \midrule  
\multirow{2}{*}{Non-Contrastive} 
  & $\Lcal_{V}$ & 200 & 2  &   69.2  & 59.9   & 37.8 &  33.1  \\
  & $\Lcal_{V}\!+\!\Lcal_{R}$ & 200 & 2    &  69.9 {\bf\textcolor{emerald!80}{(+0.7)}}  &  61.7 {\bf\textcolor{emerald!80}{(+1.8)}}  &  38.0 {\bf\textcolor{emerald!80}{(+0.2)}}   & 33.2 {\bf\textcolor{emerald!80}{(+0.1)}} \\ 
    \bottomrule
  \end{tabu}
  }
  \vspace{-2mm}
  \caption{Comparison between contrastive and non-contrastive region-matching tasks.}
  \label{table:region_matching}  
\vspace{-6mm}
\end{table}

\vspace{-2mm}
\subsection{Qualitative Studies}
\vspace{-2mm}

\begin{figure*}[t!]
	\vspace{-0mm}\centering
	\begin{tabular}{c c c}
		\hspace{-2mm}
		\includegraphics[height=2.3cm]{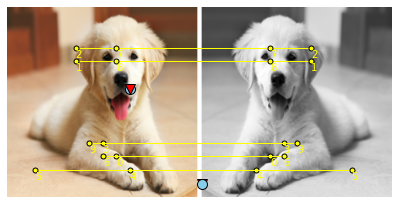}  & 
		\hspace{-3mm}
		\includegraphics[height=2.2cm]{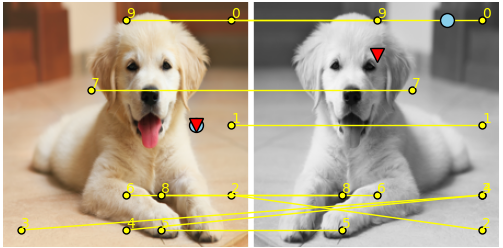} & 
		\hspace{-3mm}
		\includegraphics[height=2.2cm]{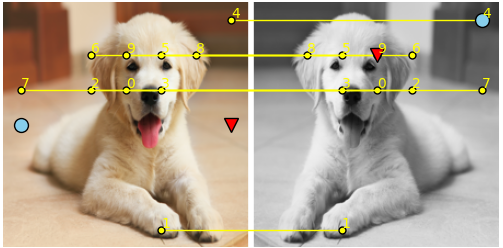} 		
		\\
		(a) DINO: DeiT-S  & 
		\hspace{-2mm}
		(b) \shortname{}: $\Lcal_V$    \vspace{2mm} & 
		\hspace{-2mm}		
		(c) \shortname{}: $\Lcal_V\!+\!\Lcal_R$   \hspace{-0mm} \\ 
	\end{tabular}
	\vspace{-3mm}
	\caption{The learned correspondences.  {\bf \textcolor{yellow!80!black}{Yellow}}  lines are the top-10 correspondences between two views, where the numbers indicates the rankings of similarity scores, yellow dots with the same number are paired.
	 }
	\vspace{-3mm}
	\label{fig:correspondences}
\end{figure*}

\paragraph{Visualization of correspondences.}  
Given two views of the same image, we use the pre-trained backbone to extract the top-layer features $\zv_1$ and $\zv_2$. 
For each feature vector in $\zv_1$, we find the feature vector in $\zv_2$ that best matches it in terms of highest cosine similarity, as defined in Equation~\eqref{eq:obj_region}. In Figure~\ref{fig:correspondences}, we show the top-10 correspondences between two views for three methods. 
In Figure~\ref{fig:correspondences} (b), 
\shortname{} with $\Lcal_V$ tends to identify pairs in the background as the most matched ones (and in a wrong way in this example). This could be a valid solution to $\Lcal_V$, as the invariance in the level of aggregated global features does not necessarily induce invariances in the local region level. This is significantly alleviated with $\Lcal_R$ (shown in Figure~\ref{fig:correspondences} (c)), a task that implicitly requires local matching. 

Surprisingly, DINO is able to learn good correspondences even without the region-level matching task. To the best of our knowledge, this is a previously unreported intriguing property of self-supervised Transformers with monolithic architectures: good semantic correspondences are automatically learned. We hypothesize that features at lower layers (image patch itself in the extreme case) can directly pass to higher layers, and the former regularizes the latter to remain discriminative. Nevertheless, the proposed $\Lcal_R$ can dramatically reduce the issue, and is good remedy to rescue the loss of semantic correspondence for the multi-stage architecture. 
In Appendix, we quantitatively measures the correspondence learning ability of these SSL methods on ImageNet validation dataset, the observations are consistent: $\Lcal_R$ improves the matching accuracy from 66\% to 91\%.

\begin{figure*}[t!]
	\vspace{-0mm}\centering
	\begin{tabular}{c c c c c c c c c}
		\hspace{-3mm}
		\includegraphics[height=1.43cm]{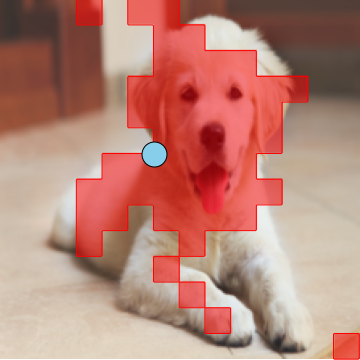}  & 
		\hspace{-4mm}
		\includegraphics[height=1.43cm]{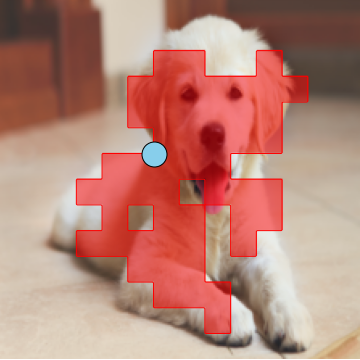}  &
		\hspace{-4mm}
		\includegraphics[height=1.43cm]{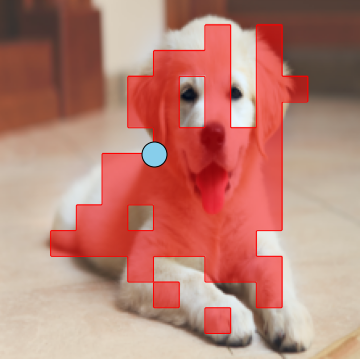}  & 
		\hspace{-2mm}
		\includegraphics[height=1.43cm]{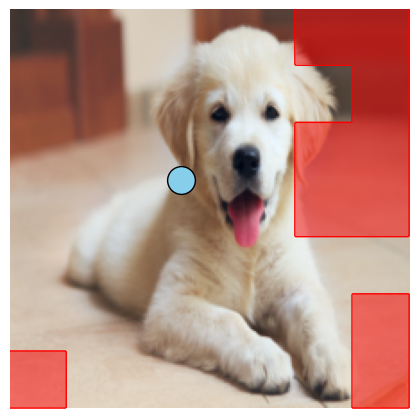} & 
		\hspace{-4mm}
		\includegraphics[height=1.43cm]{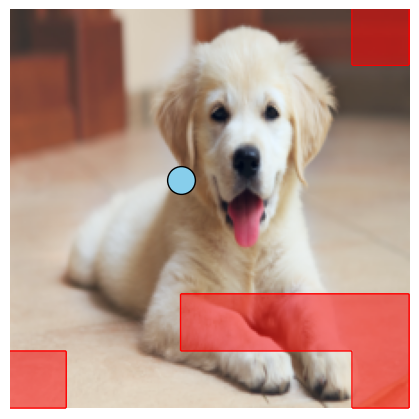} & 
		\hspace{-4mm}
		\includegraphics[height=1.43cm]{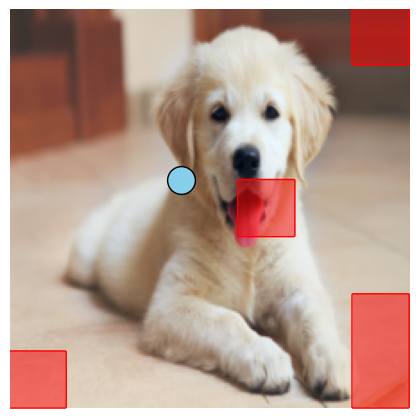} & 
		\hspace{-2mm}
		\includegraphics[height=1.43cm]{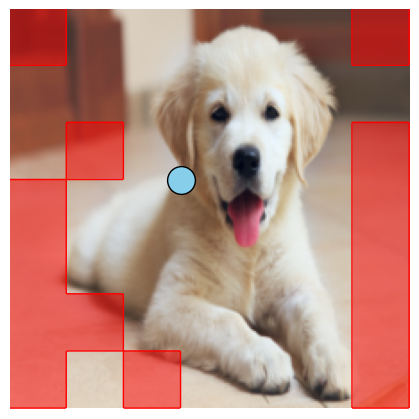} & 
		\hspace{-4mm}
		\includegraphics[height=1.43cm]{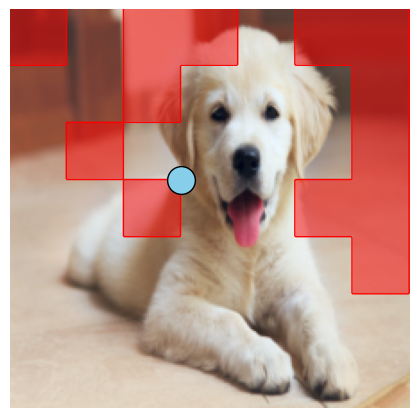} & 
		\hspace{-4mm}
		\includegraphics[height=1.43cm]{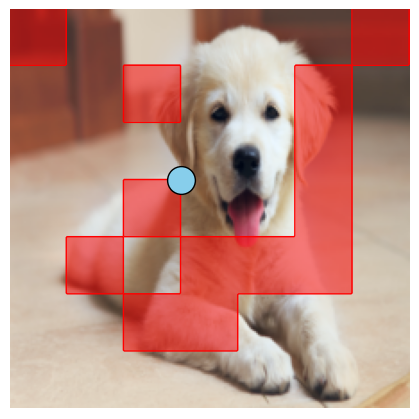} 
		\\
		\hspace{-3mm}
		\includegraphics[height=1.43cm]{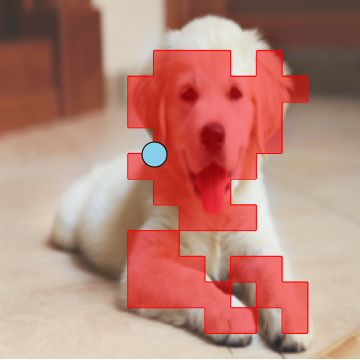}  & 
		\hspace{-4mm}
		\includegraphics[height=1.43cm]{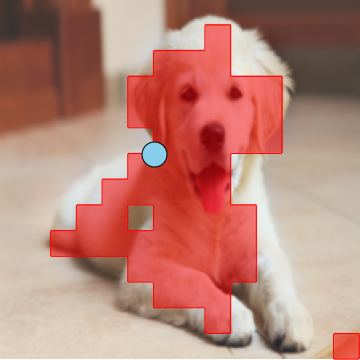}  &
		\hspace{-4mm}
		\includegraphics[height=1.43cm]{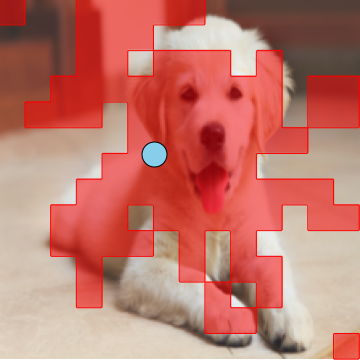}  & 
		\hspace{-2mm}
		\includegraphics[height=1.43cm]{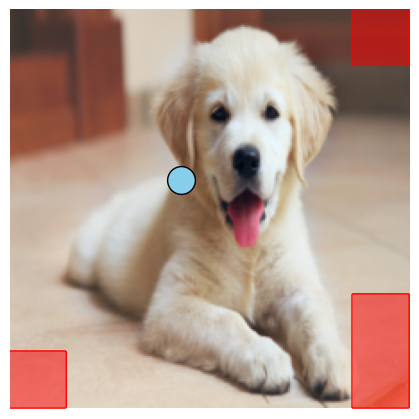} & 
		\hspace{-4mm}
		\includegraphics[height=1.43cm]{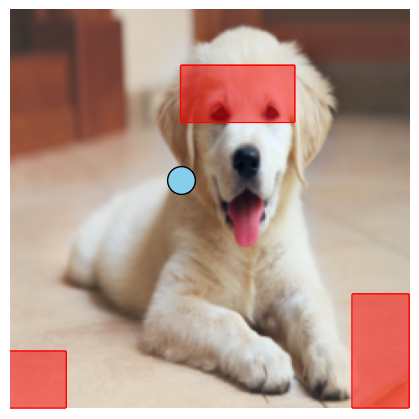} & 
		\hspace{-4mm}
		\includegraphics[height=1.43cm]{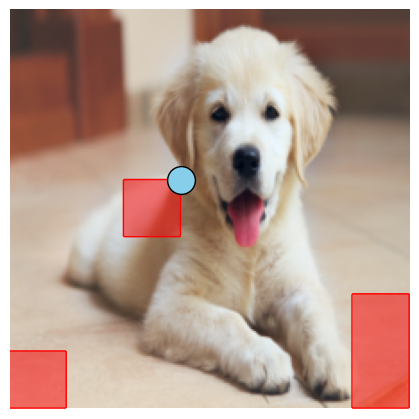} & 
		\hspace{-2mm}
		\includegraphics[height=1.43cm]{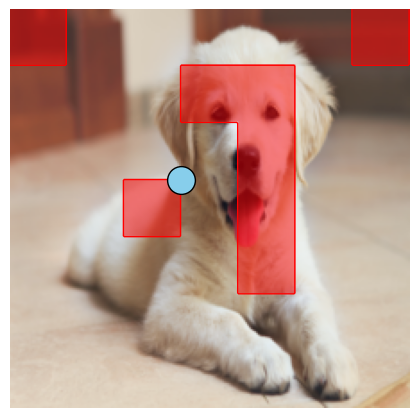} & 
		\hspace{-4mm}
		\includegraphics[height=1.43cm]{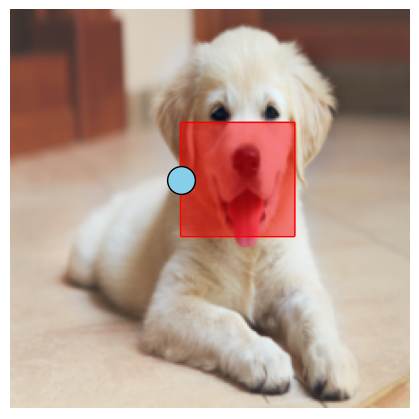} & 
		\hspace{-4mm}
		\includegraphics[height=1.43cm]{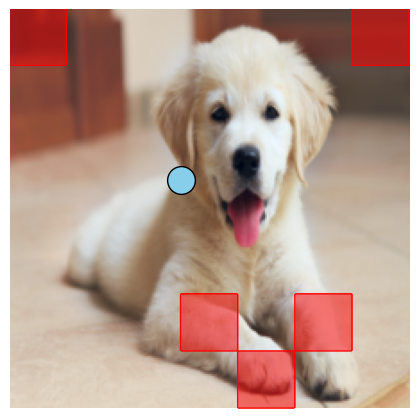}
		\\	
		\multicolumn{3}{c}{(a) DINO: DeiT-S} & 
		\multicolumn{3}{c}{(b) \shortname{}: $\Lcal_V$ } & 
		\multicolumn{3}{c}{(c) \shortname{}: $\Lcal_V\!+\!\Lcal_R$ } \\
	\end{tabular}
	\vspace{-3mm}
	\caption{Visualization of the the learned attention map for different heads in the last layer. The query is the {\bf \textcolor{blue!50}{blue}} dot in the center of the images. We visualize masks (as {\bf \textcolor{red}{red}}) obtained by thresholding the self-attention maps to
keep 60\% of the probability mass.  Note that all 6 heads are visualized for DINO with DeiT-S, and 6 out of 24 heads in \shortname{} are chosen to visualize (ranked by entropy values). Please see enlarged pictures with all heads in Appendix.
	 }
	\vspace{-5mm}
	\label{fig:attn_vis}
\end{figure*}


%

\paragraph{Visualization of attention maps.} We look at the self-attention in the different heads of the last layer in Figure~\ref{fig:attn_vis}. A local region on the edge of the main object is employed as query, and the attended regions are highlighted in red for those the query's top 60\% mass are assigned. In Appendix, we visualize more examples with different query positions.
DINO tends to automatically learn class-specific attention maps leading to foreground object segmentation, regardless of its query located in foreground or background. This is probably because main objects remain as the major invariance factor in different augmented views.
This property is lost when a multi-stage architecture is employed, as shown in \shortname{} with $\Lcal_V$. These patterns are consistent for different heads. After introducing $\Lcal_R$ for \shortname{}, we note that the attention maps become more diverse in different heads, \ie entropy values of attentions get more skewed, and attended regions are more different. This is perhaps because  $\Lcal_R$  requires each region to consider many matching tasks to regions in different augmented views, each head automatically learns to distribute the tasks and complete a few of them.







\vspace{-3mm}
\section{Conclusions}
\vspace{-3mm}
In this paper, we first discover the automatic correspondence learning property of self-supervised monolithic Transformers. Inspired by this, we present efficient self-supervised vision Transformers (\shortname{}) to with two major insights: a multi-stage Transformer architecture with sparse self-attentions, and a non-contrastive region-matching pre-training task. The synergy of both helps \shortname{} reach the SoTA performance of SSL vision systems with significantly less compute and smaller model size. 
Our study also reveals that exploration of effective solutions to learn from larger and less curated pre-training data in the wild is a key but less studied factor in paving the way toward the scaling success of SSL vision systems.


\section*{Ethics Statement}
Though self-supervised learning (SSL) has great potentials to learn powerful representation without human annotation, the existing techniques to build SoTA SSL vision systems tend to be {\bf\textcolor{red!100}{Red AI}}~\citep{schwartz2020green}: it could be environmentally unfriendly and the computational cost is extensively high. The required training resource is typically not accessible for a lab environment (thus raising barriers to participation in AI research). For example, the prior art MoCo-v3 has greatly pushes the performance limit of SSL system~\citep{chen2021empirical}. The authors kindly reported that ``it (MoCo-v3, ViT-H) takes 9.8 hours per 100 epochs using 512 TPUs. This is a gigantic scale of training: for the 300-epoch ViT-H, this amounts to $\sim$625 TPU days, or $\sim$1.7 TPU years of training.'' The SoTA model MoCo-v3 with ViT-BN-L/7 should have a higher cost than this. Even for a smaller model ViT-B, ``it takes 24 hours in 128 GPUs (vs. 2.1 hours in 256 TPUs)''. Hence, improving the efficiency of building SoTA SSL systems is of high value for the community and society to achieve {\bf\textcolor{emerald!100}{Green AI}}~\citep{schwartz2020green}.

To this end, we propose \shortname{} to provide more affordable and efficient solutions for the community to experiment and explore the directions of SoTA SSL in computer vision. Our EsViT model shows the best ImageNet linear probe performance compared with all existing SSL vision systems, and is 3.5$\times$ parameter-efficient and has 10$\times$ higher throughput than previous SoTA. This efficiency gain can significantly decrease its carbon footprint and increase its inclusivity, encouraging more researchers to participate the study of the SSL topic.

\section*{Reproducibility Statement}
Our paper provides comprehensive empirical studies on the \shortname{} algorithm. 
We provide PyTorch-style pseudo-code in Appendix. We also include an example code with instruction as supplementary material to ensure the reproducibility. 
For empirical results on both various network architecture and large-scale datasets, we provide detailed hyper-parameter specifications.
We release the pre-trained checkpoints and codebase for the research community for reproducible research.

\bibliography{egbib}

\begin{thebibliography}{83}
\providecommand{\natexlab}[1]{#1}
\providecommand{\url}[1]{\texttt{#1}}
\expandafter\ifx\csname urlstyle\endcsname\relax
  \providecommand{\doi}[1]{doi: #1}\else
  \providecommand{\doi}{doi: \begingroup \urlstyle{rm}\Url}\fi

\bibitem[Bachman et~al.(2019)Bachman, Hjelm, and
  Buchwalter]{bachman2019learning}
Philip Bachman, R~Devon Hjelm, and William Buchwalter.
\newblock Learning representations by maximizing mutual information across
  views.
\newblock In \emph{NeurIPS}, 2019.

\bibitem[Brown et~al.(2020)Brown, Mann, Ryder, Subbiah, Kaplan, Dhariwal,
  Neelakantan, Shyam, Sastry, Askell, et~al.]{brown2020language}
Tom~B Brown, Benjamin Mann, Nick Ryder, Melanie Subbiah, Jared Kaplan, Prafulla
  Dhariwal, Arvind Neelakantan, Pranav Shyam, Girish Sastry, Amanda Askell,
  et~al.
\newblock Language models are few-shot learners.
\newblock \emph{arXiv preprint arXiv:2005.14165}, 2020.

\bibitem[Carion et~al.(2020)Carion, Massa, Synnaeve, Usunier, Kirillov, and
  Zagoruyko]{carion2020end}
Nicolas Carion, Francisco Massa, Gabriel Synnaeve, Nicolas Usunier, Alexander
  Kirillov, and Sergey Zagoruyko.
\newblock End-to-end object detection with transformers.
\newblock In \emph{ECCV}, 2020.

\bibitem[Caron et~al.(2018)Caron, Bojanowski, Joulin, and
  Douze]{caron2018deepcluster}
Mathilde Caron, Piotr Bojanowski, Armand Joulin, and Matthijs Douze.
\newblock Deep clustering for unsupervised learning of visual features.
\newblock In \emph{ECCV}, 2018.

\bibitem[Caron et~al.(2020)Caron, Misra, Mairal, Goyal, Bojanowski, and
  Joulin]{caron2020unsupervised}
Mathilde Caron, Ishan Misra, Julien Mairal, Priya Goyal, Piotr Bojanowski, and
  Armand Joulin.
\newblock Unsupervised learning of visual features by contrasting cluster
  assignments.
\newblock \emph{arXiv preprint arXiv:2006.09882}, 2020.

\bibitem[Caron et~al.(2021)Caron, Touvron, Misra, J{\'e}gou, Mairal,
  Bojanowski, and Joulin]{caron2021emerging}
Mathilde Caron, Hugo Touvron, Ishan Misra, Herv{\'e} J{\'e}gou, Julien Mairal,
  Piotr Bojanowski, and Armand Joulin.
\newblock Emerging properties in self-supervised vision transformers.
\newblock \emph{arXiv preprint arXiv:2104.14294}, 2021.

\bibitem[Chen et~al.(2020{\natexlab{a}})Chen, Wang, Guo, Xu, Deng, Liu, Ma, Xu,
  Xu, and Gao]{chen2020pre}
Hanting Chen, Yunhe Wang, Tianyu Guo, Chang Xu, Yiping Deng, Zhenhua Liu, Siwei
  Ma, Chunjing Xu, Chao Xu, and Wen Gao.
\newblock Pre-trained image processing transformer.
\newblock \emph{arXiv preprint arXiv:2012.00364}, 2020{\natexlab{a}}.

\bibitem[Chen et~al.(2020{\natexlab{b}})Chen, Radford, Child, Wu, Jun,
  Dhariwal, Luan, and Sutskever]{chen2020generative}
Mark Chen, Alec Radford, Rewon Child, Jeff Wu, Heewoo Jun, Prafulla Dhariwal,
  David Luan, and Ilya Sutskever.
\newblock Generative pretraining from pixels.
\newblock In \emph{ICML}, 2020{\natexlab{b}}.

\bibitem[Chen et~al.(2020{\natexlab{c}})Chen, Kornblith, Norouzi, and
  Hinton]{chen2020simple}
Ting Chen, Simon Kornblith, Mohammad Norouzi, and Geoffrey Hinton.
\newblock A simple framework for contrastive learning of visual
  representations.
\newblock \emph{ICML}, 2020{\natexlab{c}}.

\bibitem[Chen et~al.(2020{\natexlab{d}})Chen, Kornblith, Swersky, Norouzi, and
  Hinton]{chen2020big}
Ting Chen, Simon Kornblith, Kevin Swersky, Mohammad Norouzi, and Geoffrey
  Hinton.
\newblock Big self-supervised models are strong semi-supervised learners.
\newblock \emph{arXiv preprint arXiv:2006.10029}, 2020{\natexlab{d}}.

\bibitem[Chen et~al.(2020{\natexlab{e}})Chen, Fan, Girshick, and
  He]{chen2020improved}
Xinlei Chen, Haoqi Fan, Ross Girshick, and Kaiming He.
\newblock Improved baselines with momentum contrastive learning.
\newblock \emph{arXiv preprint arXiv:2003.04297}, 2020{\natexlab{e}}.

\bibitem[Chen et~al.(2021)Chen, Xie, and He]{chen2021empirical}
Xinlei Chen, Saining Xie, and Kaiming He.
\newblock An empirical study of training self-supervised visual transformers.
\newblock \emph{arXiv preprint arXiv:2104.02057}, 2021.

\bibitem[Chen et~al.(2019)Chen, Li, Yu, Kholy, Ahmed, Gan, Cheng, and
  Liu]{chen2019uniter}
Yen-Chun Chen, Linjie Li, Licheng Yu, Ahmed~El Kholy, Faisal Ahmed, Zhe Gan,
  Yu~Cheng, and Jingjing Liu.
\newblock Uniter: Learning universal image-text representations.
\newblock \emph{arXiv preprint arXiv:1909.11740}, 2019.

\bibitem[Cuturi(2013)]{cuturi2013sinkhorn}
Marco Cuturi.
\newblock Sinkhorn distances: Lightspeed computation of optimal transport.
\newblock \emph{Advances in neural information processing systems}, 2013.

\bibitem[Dai et~al.(2020)Dai, Cai, Lin, and Chen]{dai2020up}
Zhigang Dai, Bolun Cai, Yugeng Lin, and Junying Chen.
\newblock {UP}-{DETR}: Unsupervised pre-training for object detection with
  transformers.
\newblock \emph{arXiv preprint arXiv:2011.09094}, 2020.

\bibitem[Deng et~al.(2009)Deng, Dong, Socher, Li, Li, and
  Fei-Fei]{deng2009imagenet}
Jia Deng, Wei Dong, Richard Socher, Li-Jia Li, Kai Li, and Li~Fei-Fei.
\newblock Imagenet: A large-scale hierarchical image database.
\newblock In \emph{CVPR}, 2009.

\bibitem[Devlin et~al.(2019)Devlin, Chang, Lee, and Toutanova]{devlin2019bert}
Jacob Devlin, Ming-Wei Chang, Kenton Lee, and Kristina Toutanova.
\newblock {BERT}: Pre-training of deep bidirectional transformers for language
  understanding.
\newblock \emph{NAACL}, 2019.

\bibitem[Doersch et~al.(2015)Doersch, Gupta, and
  Efros]{doersch2015unsupervised}
Carl Doersch, Abhinav Gupta, and Alexei~A Efros.
\newblock Unsupervised visual representation learning by context prediction.
\newblock In \emph{ICCV}, 2015.

\bibitem[Donahue \& Simonyan(2019)Donahue and Simonyan]{donahue2019large}
Jeff Donahue and Karen Simonyan.
\newblock Large scale adversarial representation learning.
\newblock In \emph{NeurIPS}, 2019.

\bibitem[Dosovitskiy et~al.(2015)Dosovitskiy, Fischer, Springenberg,
  Riedmiller, and Brox]{dosovitskiy2015discriminative}
Alexey Dosovitskiy, Philipp Fischer, Jost~Tobias Springenberg, Martin
  Riedmiller, and Thomas Brox.
\newblock Discriminative unsupervised feature learning with exemplar
  convolutional neural networks.
\newblock \emph{T-PAMI}, 2015.

\bibitem[Dosovitskiy et~al.(2021)Dosovitskiy, Beyer, Kolesnikov, Weissenborn,
  Zhai, Unterthiner, Dehghani, Minderer, Heigold, Gelly,
  et~al.]{dosovitskiy2020image}
Alexey Dosovitskiy, Lucas Beyer, Alexander Kolesnikov, Dirk Weissenborn,
  Xiaohua Zhai, Thomas Unterthiner, Mostafa Dehghani, Matthias Minderer, Georg
  Heigold, Sylvain Gelly, et~al.
\newblock An image is worth 16x16 words: Transformers for image recognition at
  scale.
\newblock \emph{ICLR}, 2021.

\bibitem[Frankle et~al.(2020)Frankle, Schwab, and Morcos]{frankle2020training}
Jonathan Frankle, David~J Schwab, and Ari~S Morcos.
\newblock Training {B}atchnorm and only {B}atchnorm: On the expressive power of
  random features in {CNN}s.
\newblock \emph{arXiv preprint arXiv:2003.00152}, 2020.

\bibitem[Gidaris et~al.(2018)Gidaris, Singh, and
  Komodakis]{gidaris2018unsupervised}
Spyros Gidaris, Praveer Singh, and Nikos Komodakis.
\newblock Unsupervised representation learning by predicting image rotations.
\newblock \emph{arXiv preprint arXiv:1803.07728}, 2018.

\bibitem[Goyal et~al.(2017)Goyal, Doll{\'a}r, Girshick, Noordhuis, Wesolowski,
  Kyrola, Tulloch, Jia, and He]{goyal2017accurate}
Priya Goyal, Piotr Doll{\'a}r, Ross Girshick, Pieter Noordhuis, Lukasz
  Wesolowski, Aapo Kyrola, Andrew Tulloch, Yangqing Jia, and Kaiming He.
\newblock Accurate, large minibatch {SGD}: Training {I}mage{N}et in 1 hour.
\newblock \emph{arXiv preprint arXiv:1706.02677}, 2017.

\bibitem[Goyal et~al.(2021)Goyal, Caron, Lefaudeux, Xu, Wang, Pai, Singh,
  Liptchinsky, Misra, Joulin, et~al.]{goyal2021self}
Priya Goyal, Mathilde Caron, Benjamin Lefaudeux, Min Xu, Pengchao Wang, Vivek
  Pai, Mannat Singh, Vitaliy Liptchinsky, Ishan Misra, Armand Joulin, et~al.
\newblock Self-supervised pretraining of visual features in the wild.
\newblock \emph{arXiv preprint arXiv:2103.01988}, 2021.

\bibitem[Grill et~al.(2020)Grill, Strub, Altch{\'e}, Tallec, Richemond,
  Buchatskaya, Doersch, Pires, Guo, Azar, et~al.]{grill2020bootstrap}
Jean-Bastien Grill, Florian Strub, Florent Altch{\'e}, Corentin Tallec,
  Pierre~H Richemond, Elena Buchatskaya, Carl Doersch, Bernardo~Avila Pires,
  Zhaohan~Daniel Guo, Mohammad~Gheshlaghi Azar, et~al.
\newblock Bootstrap your own latent: A new approach to self-supervised
  learning.
\newblock \emph{arXiv preprint arXiv:2006.07733}, 2020.

\bibitem[He et~al.(2016)He, Zhang, Ren, and Sun]{he2016deep}
Kaiming He, Xiangyu Zhang, Shaoqing Ren, and Jian Sun.
\newblock Deep residual learning for image recognition.
\newblock In \emph{CVPR}, 2016.

\bibitem[He et~al.(2020)He, Fan, Wu, Xie, and Girshick]{he2020momentum}
Kaiming He, Haoqi Fan, Yuxin Wu, Saining Xie, and Ross Girshick.
\newblock Momentum contrast for unsupervised visual representation learning.
\newblock In \emph{CVPR}, 2020.

\bibitem[Hjelm et~al.(2018)Hjelm, Fedorov, Lavoie-Marchildon, Grewal, Bachman,
  Trischler, and Bengio]{hjelm2018learning}
R~Devon Hjelm, Alex Fedorov, Samuel Lavoie-Marchildon, Karan Grewal, Phil
  Bachman, Adam Trischler, and Yoshua Bengio.
\newblock Learning deep representations by mutual information estimation and
  maximization.
\newblock \emph{arXiv preprint arXiv:1808.06670}, 2018.

\bibitem[Ji et~al.(2019)Ji, Henriques, and Vedaldi]{ji2019invariant}
Xu~Ji, Jo{\~a}o~F Henriques, and Andrea Vedaldi.
\newblock Invariant information clustering for unsupervised image
  classification and segmentation.
\newblock In \emph{ICCV}, 2019.

\bibitem[Kuznetsova et~al.(2020)Kuznetsova, Rom, Alldrin, Uijlings, Krasin,
  Pont-Tuset, Kamali, Popov, Malloci, Kolesnikov, et~al.]{kuznetsova2020open}
Alina Kuznetsova, Hassan Rom, Neil Alldrin, Jasper Uijlings, Ivan Krasin, Jordi
  Pont-Tuset, Shahab Kamali, Stefan Popov, Matteo Malloci, Alexander
  Kolesnikov, et~al.
\newblock The open images dataset v4.
\newblock \emph{International Journal of Computer Vision}, 2020.

\bibitem[Larsson et~al.(2016)Larsson, Maire, and
  Shakhnarovich]{larsson2016learning}
Gustav Larsson, Michael Maire, and Gregory Shakhnarovich.
\newblock Learning representations for automatic colorization.
\newblock In \emph{ECCV}, 2016.

\bibitem[Li et~al.(2020{\natexlab{a}})Li, Li, Zhang, Peng, Zhou, and
  Gao]{li2020self}
Chunyuan Li, Xiujun Li, Lei Zhang, Baolin Peng, Mingyuan Zhou, and Jianfeng
  Gao.
\newblock Self-supervised pre-training with hard examples improves visual
  representations.
\newblock \emph{arXiv preprint arXiv:2012.13493}, 2020{\natexlab{a}}.

\bibitem[Li et~al.(2019{\natexlab{a}})Li, Duan, Fang, Jiang, and
  Zhou]{li2019unicoder}
Gen Li, Nan Duan, Yuejian Fang, Daxin Jiang, and Ming Zhou.
\newblock Unicoder-{VL}: A universal encoder for vision and language by
  cross-modal pre-training.
\newblock \emph{arXiv preprint arXiv:1908.06066}, 2019{\natexlab{a}}.

\bibitem[Li et~al.(2020{\natexlab{b}})Li, Zhou, Xiong, Socher, and
  Hoi]{li2020prototypical}
Junnan Li, Pan Zhou, Caiming Xiong, Richard Socher, and Steven~CH Hoi.
\newblock Prototypical contrastive learning of unsupervised representations.
\newblock \emph{arXiv preprint arXiv:2005.04966}, 2020{\natexlab{b}}.

\bibitem[Li et~al.(2019{\natexlab{b}})Li, Yatskar, Yin, Hsieh, and
  Chang]{li2019visualbert}
Liunian~Harold Li, Mark Yatskar, Da~Yin, Cho-Jui Hsieh, and Kai-Wei Chang.
\newblock Visualbert: {A} simple and performant baseline for vision and
  language.
\newblock \emph{arXiv preprint arXiv:1908.03557}, 2019{\natexlab{b}}.

\bibitem[Li et~al.(2017)Li, Wang, Li, Agustsson, and Van~Gool]{li2017webvision}
Wen Li, Limin Wang, Wei Li, Eirikur Agustsson, and Luc Van~Gool.
\newblock Webvision database: Visual learning and understanding from web data.
\newblock \emph{arXiv preprint arXiv:1708.02862}, 2017.

\bibitem[Li et~al.(2019{\natexlab{c}})Li, Wang, Hu, and Yang]{li2019selective}
Xiang Li, Wenhai Wang, Xiaolin Hu, and Jian Yang.
\newblock Selective kernel networks.
\newblock In \emph{CVPR}, 2019{\natexlab{c}}.

\bibitem[Li et~al.(2020{\natexlab{c}})Li, Yin, Li, Zhang, Hu, Zhang, Wang, Hu,
  Dong, Wei, Choi, and Gao]{li2020oscar}
Xiujun Li, Xi~Yin, Chunyuan Li, Pengchuan Zhang, Xiaowei Hu, Lei Zhang, Lijuan
  Wang, Houdong Hu, Li~Dong, Furu Wei, Yejin Choi, and Jianfeng Gao.
\newblock Oscar: Object-semantics aligned pre-training for vision-language
  tasks.
\newblock In \emph{ECCV}, 2020{\natexlab{c}}.

\bibitem[Liu et~al.(2021)Liu, Lin, Cao, Hu, Wei, Zhang, Lin, and
  Guo]{liu2021Swin}
Ze~Liu, Yutong Lin, Yue Cao, Han Hu, Yixuan Wei, Zheng Zhang, Stephen Lin, and
  Baining Guo.
\newblock Swin transformer: Hierarchical vision transformer using shifted
  windows.
\newblock \emph{arXiv preprint arXiv:2103.14030}, 2021.

\bibitem[Loshchilov \& Hutter(2018)Loshchilov and Hutter]{loshchilov2018fixing}
Ilya Loshchilov and Frank Hutter.
\newblock Fixing weight decay regularization in {A}dam.
\newblock \emph{arXiv preprint arXiv:1706.02677}, 2018.

\bibitem[Lu et~al.(2019)Lu, Batra, Parikh, and Lee]{lu2019vilbert}
Jiasen Lu, Dhruv Batra, Devi Parikh, and Stefan Lee.
\newblock Vil{BERT}: Pretraining task-agnostic visiolinguistic representations
  for vision-and-language tasks.
\newblock \emph{NeurIPS}, 2019.

\bibitem[Misra \& Maaten(2020)Misra and Maaten]{misra2020self}
Ishan Misra and Laurens van~der Maaten.
\newblock Self-supervised learning of pretext-invariant representations.
\newblock In \emph{CVPR}, pp.\  6707--6717, 2020.

\bibitem[Noroozi \& Favaro(2016)Noroozi and Favaro]{noroozi2016unsupervised}
Mehdi Noroozi and Paolo Favaro.
\newblock Unsupervised learning of visual representations by solving jigsaw
  puzzles.
\newblock In \emph{ECCV}, 2016.

\bibitem[Oord et~al.(2018)Oord, Li, and Vinyals]{oord2018representation}
Aaron van~den Oord, Yazhe Li, and Oriol Vinyals.
\newblock Representation learning with contrastive predictive coding.
\newblock \emph{arXiv preprint arXiv:1807.03748}, 2018.

\bibitem[Parmar et~al.(2018)Parmar, Vaswani, Uszkoreit, Kaiser, Shazeer, Ku,
  and Tran]{parmar2018image}
Niki Parmar, Ashish Vaswani, Jakob Uszkoreit, Lukasz Kaiser, Noam Shazeer,
  Alexander Ku, and Dustin Tran.
\newblock Image transformer.
\newblock In \emph{International Conference on Machine Learning}, 2018.

\bibitem[Pathak et~al.(2016)Pathak, Krahenbuhl, Donahue, Darrell, and
  Efros]{pathak2016context}
Deepak Pathak, Philipp Krahenbuhl, Jeff Donahue, Trevor Darrell, and Alexei~A
  Efros.
\newblock Context encoders: Feature learning by inpainting.
\newblock In \emph{CVPR}, 2016.

\bibitem[Pu et~al.(2016)Pu, Gan, Henao, Yuan, Li, Stevens, and
  Carin]{pu2016variational}
Yunchen Pu, Zhe Gan, Ricardo Henao, Xin Yuan, Chunyuan Li, Andrew Stevens, and
  Lawrence Carin.
\newblock Variational autoencoder for deep learning of images, labels and
  captions.
\newblock \emph{NIPS}, 2016.

\bibitem[Radford et~al.(2018)Radford, Narasimhan, Salimans, and
  Sutskever]{radford2018improving}
Alec Radford, Karthik Narasimhan, Tim Salimans, and Ilya Sutskever.
\newblock Improving language understanding by generative pre-training.
\newblock \emph{OpenAI Blog}, 2018.

\bibitem[Radford et~al.(2021)Radford, Kim, Hallacy, Ramesh, Goh, Agarwal,
  Sastry, Askell, Mishkin, Clark, et~al.]{radford2021learning}
Alec Radford, Jong~Wook Kim, Chris Hallacy, Aditya Ramesh, Gabriel Goh,
  Sandhini Agarwal, Girish Sastry, Amanda Askell, Pamela Mishkin, Jack Clark,
  et~al.
\newblock Learning transferable visual models from natural language
  supervision.
\newblock \emph{arXiv preprint arXiv:2103.00020}, 2021.

\bibitem[Ramesh et~al.(2021)Ramesh, Pavlov, Goh, Gray, Voss, Radford, Chen, and
  Sutskever]{ramesh2021zero}
Aditya Ramesh, Mikhail Pavlov, Gabriel Goh, Scott Gray, Chelsea Voss, Alec
  Radford, Mark Chen, and Ilya Sutskever.
\newblock Zero-shot text-to-image generation.
\newblock \emph{arXiv preprint arXiv:2102.12092}, 2021.

\bibitem[Schwartz et~al.(2020)Schwartz, Dodge, Smith, and
  Etzioni]{schwartz2020green}
Roy Schwartz, Jesse Dodge, Noah~A Smith, and Oren Etzioni.
\newblock Green {AI}.
\newblock \emph{Communications of the ACM}, 2020.

\bibitem[Simonyan \& Zisserman(2014)Simonyan and Zisserman]{simonyan2014very}
Karen Simonyan and Andrew Zisserman.
\newblock Very deep convolutional networks for large-scale image recognition.
\newblock \emph{arXiv preprint arXiv:1409.1556}, 2014.

\bibitem[Su et~al.(2019)Su, Zhu, Cao, Li, Lu, Wei, and Dai]{su2019vl}
Weijie Su, Xizhou Zhu, Yue Cao, Bin Li, Lewei Lu, Furu Wei, and Jifeng Dai.
\newblock {VL-BERT}: Pre-training of generic visual-linguistic representations.
\newblock \emph{arXiv preprint arXiv:1908.08530}, 2019.

\bibitem[Tan \& Bansal(2019)Tan and Bansal]{tan2019lxmert}
Hao Tan and Mohit Bansal.
\newblock {LXMERT}: Learning cross-modality encoder representations from
  transformers.
\newblock \emph{EMNLP}, 2019.

\bibitem[Tian et~al.(2020)Tian, Sun, Poole, Krishnan, Schmid, and
  Isola]{tian2020makes}
Yonglong Tian, Chen Sun, Ben Poole, Dilip Krishnan, Cordelia Schmid, and
  Phillip Isola.
\newblock What makes for good views for contrastive learning.
\newblock \emph{arXiv preprint arXiv:2005.10243}, 2020.

\bibitem[Tolstikhin et~al.(2021)Tolstikhin, Houlsby, Kolesnikov, Beyer, Zhai,
  Unterthiner, Yung, Keysers, Uszkoreit, Lucic, et~al.]{tolstikhin2021mlp}
Ilya Tolstikhin, Neil Houlsby, Alexander Kolesnikov, Lucas Beyer, Xiaohua Zhai,
  Thomas Unterthiner, Jessica Yung, Daniel Keysers, Jakob Uszkoreit, Mario
  Lucic, et~al.
\newblock {MLP}-mixer: An all-{MLP} architecture for vision.
\newblock \emph{arXiv preprint arXiv:2105.01601}, 2021.

\bibitem[Touvron et~al.(2020)Touvron, Cord, Douze, Massa, Sablayrolles, and
  J{\'e}gou]{touvron2020training}
Hugo Touvron, Matthieu Cord, Matthijs Douze, Francisco Massa, Alexandre
  Sablayrolles, and Herv{\'e} J{\'e}gou.
\newblock Training data-efficient image transformers \& distillation through
  attention.
\newblock \emph{arXiv preprint arXiv:2012.12877}, 2020.

\bibitem[Trinh et~al.(2019)Trinh, Luong, and Le]{trinh2019selfie}
Trieu~H Trinh, Minh-Thang Luong, and Quoc~V Le.
\newblock Selfie: Self-supervised pretraining for image embedding.
\newblock \emph{arXiv preprint arXiv:1906.02940}, 2019.

\bibitem[Vaswani et~al.(2017)Vaswani, Shazeer, Parmar, Uszkoreit, Jones, Gomez,
  Kaiser, and Polosukhin]{vaswani2017attention}
Ashish Vaswani, Noam Shazeer, Niki Parmar, Jakob Uszkoreit, Llion Jones,
  Aidan~N Gomez, {\L}ukasz Kaiser, and Illia Polosukhin.
\newblock Attention is all you need.
\newblock In \emph{NIPS}, 2017.

\bibitem[Vaswani et~al.(2021)Vaswani, Ramachandran, Srinivas, Parmar, Hechtman,
  and Shlens]{vaswani2021scaling}
Ashish Vaswani, Prajit Ramachandran, Aravind Srinivas, Niki Parmar, Blake
  Hechtman, and Jonathon Shlens.
\newblock Scaling local self-attention for parameter efficient visual
  backbones.
\newblock \emph{CVPR}, 2021.

\bibitem[Wang et~al.(2020{\natexlab{a}})Wang, Zhu, Adam, Yuille, and
  Chen]{wang2020max}
Huiyu Wang, Yukun Zhu, Hartwig Adam, Alan Yuille, and Liang-Chieh Chen.
\newblock Max-deeplab: End-to-end panoptic segmentation with mask transformers.
\newblock \emph{arXiv preprint arXiv:2012.00759}, 2020{\natexlab{a}}.

\bibitem[Wang et~al.(2021)Wang, Xie, Li, Fan, Song, Liang, Lu, Luo, and
  Shao]{wang2021pyramid}
Wenhai Wang, Enze Xie, Xiang Li, Deng-Ping Fan, Kaitao Song, Ding Liang, Tong
  Lu, Ping Luo, and Ling Shao.
\newblock Pyramid vision transformer: A versatile backbone for dense prediction
  without convolutions.
\newblock \emph{arXiv preprint arXiv:2102.12122}, 2021.

\bibitem[Wang et~al.(2020{\natexlab{b}})Wang, Zhang, Shen, Kong, and
  Li]{wang2020dense}
Xinlong Wang, Rufeng Zhang, Chunhua Shen, Tao Kong, and Lei Li.
\newblock Dense contrastive learning for self-supervised visual pre-training.
\newblock \emph{arXiv preprint arXiv:2011.09157}, 2020{\natexlab{b}}.

\bibitem[Wang et~al.(2020{\natexlab{c}})Wang, Xu, Wang, Shen, Cheng, Shen, and
  Xia]{wang2020end}
Yuqing Wang, Zhaoliang Xu, Xinlong Wang, Chunhua Shen, Baoshan Cheng, Hao Shen,
  and Huaxia Xia.
\newblock End-to-end video instance segmentation with transformers.
\newblock \emph{arXiv preprint arXiv:2011.14503}, 2020{\natexlab{c}}.

\bibitem[Wu et~al.(2021)Wu, Xiao, Codella, Liu, Dai, Yuan, and
  Zhang]{wu2021cvt}
Haiping Wu, Bin Xiao, Noel Codella, Mengchen Liu, Xiyang Dai, Lu~Yuan, and Lei
  Zhang.
\newblock Cvt: Introducing convolutions to vision transformers.
\newblock \emph{arXiv preprint arXiv:2103.15808}, 2021.

\bibitem[Xie et~al.(2021{\natexlab{a}})Xie, Ding, Wang, Zhan, Xu, Li, and
  Luo]{xie2021detco}
Enze Xie, Jian Ding, Wenhai Wang, Xiaohang Zhan, Hang Xu, Zhenguo Li, and Ping
  Luo.
\newblock Detco: Unsupervised contrastive learning for object detection.
\newblock \emph{arXiv preprint arXiv:2102.04803}, 2021{\natexlab{a}}.

\bibitem[Xie et~al.(2016)Xie, Girshick, and Farhadi]{xie2016unsupervised}
Junyuan Xie, Ross Girshick, and Ali Farhadi.
\newblock Unsupervised deep embedding for clustering analysis.
\newblock In \emph{ICML}, 2016.

\bibitem[Xie et~al.(2021{\natexlab{b}})Xie, Lin, Yao, Zhang, Dai, Cao, and
  Hu]{xie2021moby}
Zhenda Xie, Yutong Lin, Zhuliang Yao, Zheng Zhang, Qi~Dai, Yue Cao, and Han Hu.
\newblock Self-supervised learning with swin transformers.
\newblock \emph{arXiv preprint arXiv:2105.04553}, 2021{\natexlab{b}}.

\bibitem[Xie et~al.(2021{\natexlab{c}})Xie, Lin, Zhang, Cao, Lin, and
  Hu]{xie2021propagate}
Zhenda Xie, Yutong Lin, Zheng Zhang, Yue Cao, Stephen Lin, and Han Hu.
\newblock Propagate yourself: Exploring pixel-level consistency for
  unsupervised visual representation learning.
\newblock In \emph{Proceedings of the IEEE/CVF Conference on Computer Vision
  and Pattern Recognition}, 2021{\natexlab{c}}.

\bibitem[Xiong et~al.(2020)Xiong, Ren, and Urtasun]{xiong2020loco}
Yuwen Xiong, Mengye Ren, and Raquel Urtasun.
\newblock Loco: Local contrastive representation learning.
\newblock \emph{arXiv preprint arXiv:2008.01342}, 2020.

\bibitem[Yang et~al.(2021)Yang, Wu, Zhou, and Lin]{yang2021instance}
Ceyuan Yang, Zhirong Wu, Bolei Zhou, and Stephen Lin.
\newblock Instance localization for self-supervised detection pretraining.
\newblock In \emph{Proceedings of the IEEE/CVF Conference on Computer Vision
  and Pattern Recognition}, 2021.

\bibitem[Yang et~al.(2020)Yang, Yang, Fu, Lu, and Guo]{yang2020learning}
Fuzhi Yang, Huan Yang, Jianlong Fu, Hongtao Lu, and Baining Guo.
\newblock Learning texture transformer network for image super-resolution.
\newblock In \emph{CVPR}, 2020.

\bibitem[Yang et~al.(2016)Yang, Parikh, and Batra]{yang2016joint}
Jianwei Yang, Devi Parikh, and Dhruv Batra.
\newblock Joint unsupervised learning of deep representations and image
  clusters.
\newblock In \emph{CVPR}, 2016.

\bibitem[Yonglong et~al.(2021)Yonglong, Henaff, and van~den Oord]{tian2021dnc}
Tian Yonglong, Olivier~J. Henaff, and Aaron van~den Oord.
\newblock Divide and contrast: Self-supervised learning from uncurated data.
\newblock \emph{arXiv preprint arXiv:2105.08054}, 2021.

\bibitem[Zhan et~al.(2020)Zhan, Xie, Liu, Ong, and Loy]{zhan2020online}
Xiaohang Zhan, Jiahao Xie, Ziwei Liu, Yew-Soon Ong, and Chen~Change Loy.
\newblock Online deep clustering for unsupervised representation learning.
\newblock In \emph{CVPR}, pp.\  6688--6697, 2020.

\bibitem[Zhang et~al.(2021)Zhang, Dai, Yang, Xiao, Yuan, Zhang, and
  Gao]{zhang2021vil}
Pengchuan Zhang, Xiyang Dai, Jianwei Yang, Bin Xiao, Lu~Yuan, Lei Zhang, and
  Jianfeng Gao.
\newblock Multi-scale vision longformer: A new vision transformer for
  high-resolution image encoding.
\newblock \emph{arXiv preprint arXiv:2103.15358}, 2021.

\bibitem[Zhang et~al.(2016)Zhang, Isola, and Efros]{zhang2016colorful}
Richard Zhang, Phillip Isola, and Alexei~A Efros.
\newblock Colorful image colorization.
\newblock In \emph{ECCV}, 2016.

\bibitem[Zhang et~al.(2017)Zhang, Isola, and Efros]{zhang2017split}
Richard Zhang, Phillip Isola, and Alexei~A Efros.
\newblock Split-brain autoencoders: Unsupervised learning by cross-channel
  prediction.
\newblock In \emph{CVPR}, 2017.

\bibitem[Zheng et~al.(2020)Zheng, Gao, Wang, Li, and Dong]{zheng2020end}
Minghang Zheng, Peng Gao, Xiaogang Wang, Hongsheng Li, and Hao Dong.
\newblock End-to-end object detection with adaptive clustering transformer.
\newblock \emph{arXiv preprint arXiv:2011.09315}, 2020.

\bibitem[Zhou et~al.(2020)Zhou, Palangi, Zhang, Hu, Corso, and
  Gao]{zhou2019unified}
Luowei Zhou, Hamid Palangi, Lei Zhang, Houdong Hu, Jason~J Corso, and Jianfeng
  Gao.
\newblock Unified vision-language pre-training for image captioning and {VQA}.
\newblock \emph{AAAI}, 2020.

\bibitem[Zhu et~al.(2020)Zhu, Su, Lu, Li, Wang, and Dai]{zhu2020deformable}
Xizhou Zhu, Weijie Su, Lewei Lu, Bin Li, Xiaogang Wang, and Jifeng Dai.
\newblock Deformable detr: Deformable transformers for end-to-end object
  detection.
\newblock \emph{arXiv preprint arXiv:2010.04159}, 2020.

\bibitem[Zhuang et~al.(2019)Zhuang, Zhai, and Yamins]{zhuang2019local}
Chengxu Zhuang, Alex~Lin Zhai, and Daniel Yamins.
\newblock Local aggregation for unsupervised learning of visual embeddings.
\newblock In \emph{CVPR}, 2019.

\end{thebibliography}
\bibliographystyle{iclr2022_conference}

\newpage
\appendix

\appendix


\section{Methods}

\subsection{Algorithms}
We summarize the training algorithm procedure of \shortname{}  with $\Lcal_V\!+\!\Lcal_R$ in Algorithm~\ref{alg:esvit}. To clearly outline the main idea of the algorithm, we show the algorithm for two augmented views. For the full algorithm to deal with multi-crop, please refer to our codebase. In Algorithm~\ref{alg:esvit}, for a mini-batch of size $n$, the teacher/student network consists of three output variables:
(1) $\var{p} \in \R^{n\times K}$ is the probability vector for the view-level representation, output by an MLP head. 
(2) $\var{z} \in \R^{n \times T \times P} $ is the feature map, containing $T$ region-level features of dimension $P$. 
(3) $\var{pz} \in \R^{n \times T \times K} $ are probability vectors of $\var{z}$, output by a different MLP head.

\begin{algorithm}
  \caption{\shortname{} with $\Lcal_V\!+\!\Lcal_R$, pseudocode with 2-crop.}
  \label{alg:esvit}
  \footnotesize
  \Comment{gs, gt: student and teacher networks}
  \Comment{Cv, Cr: view and region center (K) }
  \Comment{tmp\_s, tmp\_t: student and teacher temperatures}
  \Comment{a, b: network and center momentum rates.}
  \Comment{n: batch size, K: MLP-head-projected probability vector length, \hspace{3cm} T: last layer feature map length, P: last layer feature vector length}
  $\var{gt.params} = \var{gs.params}$ \\
  \Comment{The main training loop}
    \For{ \var{x} in \var{loader}}{
       \var{x1}, \var{x2} = \FuncCall{augment}{x}, \FuncCall{augment}{x}~~~ \Comment{two random views}
       \;
       \Comment{student output, p:n$\times$K, pz:n$\times$T$\times$K, z:n$\times$T$\times$P} 
       \var{p\_s1}, \var{pz\_s1}, \var{z\_s1}  = \FuncCall{gs}{x1}  \\
       \var{p\_s2}, \var{pz\_s2}, \var{z\_s2}  = \FuncCall{gs}{x2}  \\
       \Comment{teacher output, p:n$\times$K, pz:n$\times$T$\times$K, z:n$\times$T$\times$P} 
       \var{p\_t1}, \var{pz\_t1}, \var{z\_t1}  = \FuncCall{gt}{x1}  \\
       \var{p\_t2}, \var{pz\_t2}, \var{z\_t2}  = \FuncCall{gt}{x2} 
       \;\;
      \Comment{view-level loss}
      $\var{loss\_v} = \FuncCall{Hv}{\var{p\_s1}, \var{p\_t2}}/2 + \FuncCall{Hv}{\var{p\_s2}, \var{p\_t1}}/2 $\;
      \Comment{region-level loss}
      $\var{loss\_r} = \FuncCall{Hr}{\var{pz\_s1}, \var{pz\_t2}, \var{z\_s1}, \var{z\_t2}}/2 + \FuncCall{Hr}{\var{pz\_s2}, \var{pz\_t1}, \var{z\_s2}, \var{z\_t1}}/2$\;      
      $\var{loss} = \var{loss\_v}/2 + \var{loss\_r}/2$\;
      $\var{loss.backward()}$~~~\Comment{back-propagate}  
      \;
      \Comment{update student, teacher and centers} 
      $\var{update(gs)}$~~~~~~~\hspace{6cm}\Comment{AdamW for student}  
      $\var{gt.params} = a*\var{gt.params} + (1-a)*\var{gs.params}$~~~~~~~~\Comment{EMA for teacher}  
      $\var{Cv} = b*\var{Cv} + (1-b)*\var{cat([p\_t1,p\_t2].mean(0))}$~~~~~~~\Comment{EMA for view center} 
      $\var{Cr} = b*\var{Cr} + (1-b)*\var{cat([pz\_t1,pz\_t2].mean(0))}$~~~\Comment{EMA for region center}
    }
    \;
  \Comment{The view-level loss function}
  \Function{Hv(\var{s}, \var{t})}{
    $\var{t = t.detach()}$~~~~\Comment{stop gradient}  
    $\var{s = softmax(s / tmp\_s, dim=-1)}$ \;
    $\var{t = softmax((t - Cv) / tmp\_t, dim=-1)}$ \;
    \Return{ - (t * log(s)).sum(dim=-1).mean() }\;
  }
  \Comment{The region-level loss function}
  \Function{Hr(\var{ps}, \var{pt}, \var{zs}, \var{zt})}{
    $\var{pt = pt.detach()}$~~\hspace{6cm}~~~~~~\Comment{stop gradient}  
    $\var{ps = softmax(ps / tmp\_s, dim=-1)}$ ~\hspace{3cm}~~~~~~\Comment{n$\times$T$\times$K} 
    $\var{pt = softmax((pt - Cr) / tmp\_t, dim=-1)}$ ~\hspace{2cm}~~\Comment{n$\times$T$\times$K} 
    $\var{sim\_matrix} = \FuncCall{torch.matmul}{zs , zt.permute(0, 2, 1)}$~~ \Comment{n$\times$T$\times$T} 
    $\var{sim\_idx} = \var{sim\_matrix.max(dim=-1)[1].unsqueeze(2)}$~~~~~~~~~~\Comment{n$\times$T$\times$1} 
    $\var{pt\_idxed} = \FuncCall{torch.gather}{pt, 1, sim\_idx.expand(-1, -1, pt.size(2))}$\;
    \Return{ - (pt\_idxed * log(ps)).sum(dim=-1).mean() }\;
  }
\end{algorithm}

\vspace{-2mm}
\subsection{Network architecture configurations and implementation details}
\vspace{-2mm}
Inspired by great successes of the multi-stage ConvNet architecture such as VGG~\citep{simonyan2014very}/ResNets~\citep{he2016deep} for computer vision, the multi-stage Transformer-based networks have been explored very recently in the supervised learning setting~\citep{vaswani2021scaling,wang2021pyramid,liu2021Swin,zhang2021vil,wu2021cvt}. 
%
In multi-stage vision Transformers, since a larger number of patches is often produced at the early stages, an efficient Transformer with sparse self-attentions is considered to reduce the computational complexity. The basic idea is to split the feature maps into non-overlapping local windows (with size $W\!\times \!W$), and self-attention is performed within each local window. This however has one drawback that features in different local windows cannot interact. Various methods have been proposed to best approximate full-attention, with different trade-off between accuracy and efficiency.

We briefly describe three schemes as follows, and benchmark them in the experiments.
$(\RN{1})$ 
{\em Swin Transformer}~\citep{liu2021Swin}: A shifted window partitioning approach is proposed, which alternates between two partitioning configurations in consecutive Transformer blocks, so that each local feature is grouped into different windows in self-attentions.
$(\RN{2})$
{\em Vision Longformer (ViL)}~\citep{zhang2021vil}: Features in each local window are further allowed to attend all features in the 8-neighboring windows.
$(\RN{3})$
{\em  Convolution vision Transformer (CvT)}~\citep{wu2021cvt}: Features in neighboring windows are considered in the convolutional projection in self-attentions.

The window size is set to $W=7$ by default. The query dimension of each head in self-attentions is $d = 32$, and the hidden layer width of
each MLP is $4\times$ of its input's width, for all experiments. The architecture configurations of model variants employed in the experiments are summarized in Table~\ref{tab:model_config}. Some notable implementation detailed are described as follows:

\begin{minipage}{0.99\textwidth}
\centering

\begin{itemize}[leftmargin=5.5mm]
\item The three configurations Swin-T, Swin-S and Swin-B indicate Tiny, Small, and Base models, respectively, which are almost identical to the original implementation~\citep{liu2021Swin}, except that we add special treatments to deal with input augmented views of different resolutions, when the resolution (feature map size more specifically) is not divisible by the window size (\ie resolution 96 and window size=7 or 14).
\item Swin-T and Swin-S with window size $W\!=\!14$ are customized by us to allow full self-attention in stage 3 (where the majority of model capacity is allocated to) and stage 4, to study the impact of longer sequences in \shortname{}.
\item In the original ViL~\citep{zhang2021vil} and CvT~\citep{wu2021cvt} papers, different positional embedding strategies and multi-stage network configurations were employed. We modify them by only utilizing relative position bias and their proposed sparse self-attention mechanisms, and create a similar 4-stage architecture with Swin-T for fair comparison.
\end{itemize}
\end{minipage}

\paragraph{Relative Position Bias.} To facilitate SSL, we consider relative position bias~\citep{liu2021Swin} to characterize the spatial information between features for the three efficient Transformers aforementioned, and do not use absolute position embeddings. 
This is because augmented views of varied resolutions can be cropped from anywhere in an image in SSL,  maintaining the relative positions is easy in implementation, and is largely sufficient for invariance learning among these views.

\begin{figure*}[t!]
	\vspace{-0mm}\centering
	\begin{tabular}{c c}
		\hspace{-4mm}
		\includegraphics[height=5.1cm]{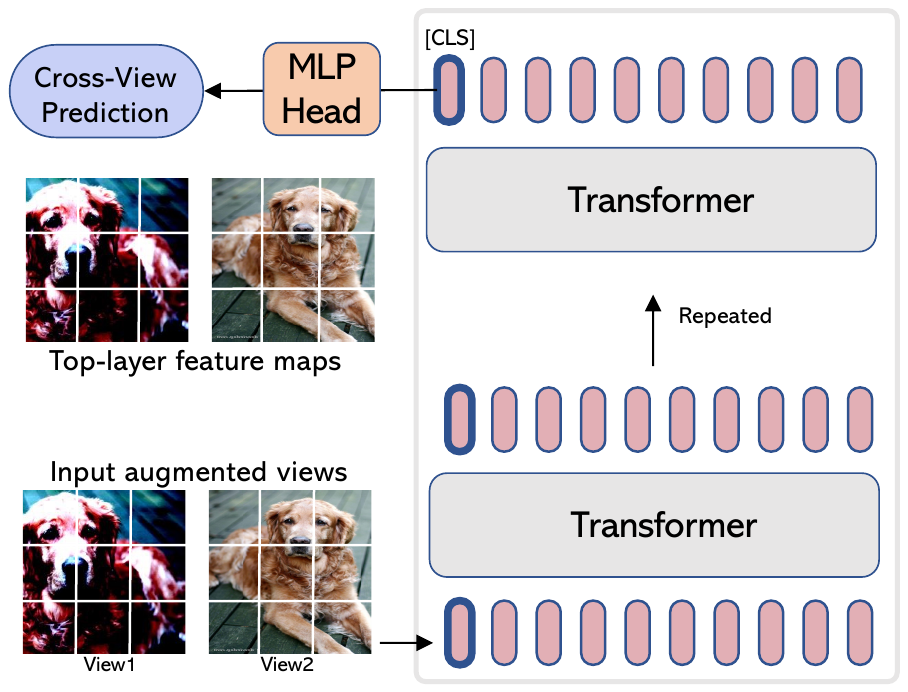}  & 
		\hspace{-0mm}
		\includegraphics[height=5.1cm]{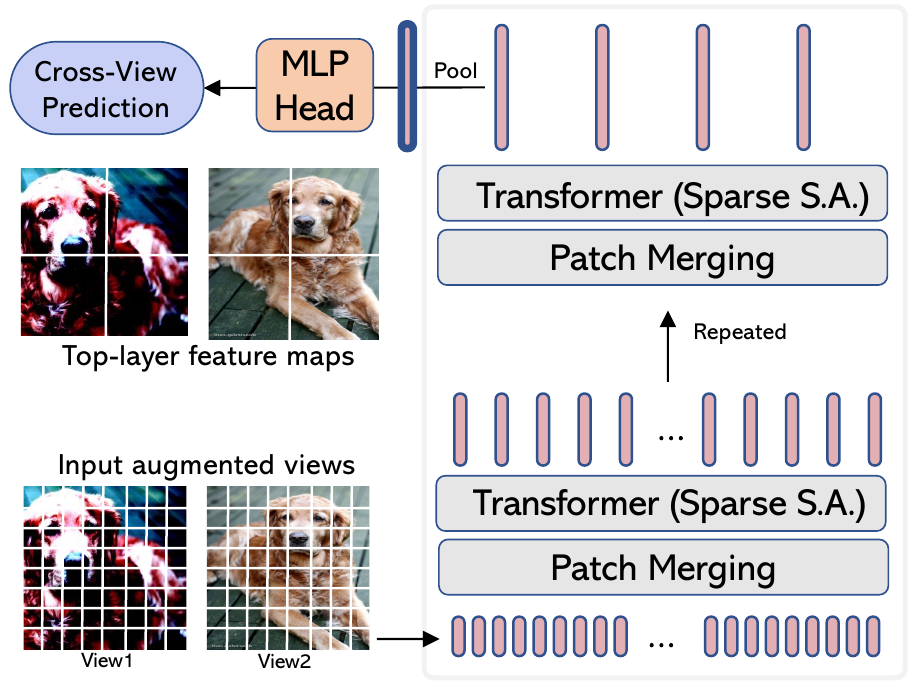} \\
		(a) Baseline monolithic architecture \vspace{2mm} & 
		\hspace{-6mm}
		(b) Multi-stage architecture \hspace{-0mm} \\ 
	\end{tabular}
	\vspace{-3mm}
	\caption{Architecture comparison. (a) The monolithic transformer. For all layers, the transformer blocks share the same network configurations and input token sequence sizes are the same. (b) The multi-stage Transformer organizes an input image into a long sequence of smaller patches, sparse self-attentions (S.A.) are utilized at early stages to maintain model expressiveness while reducing computational complexity; The neighboring tokens at an intermediate layer are gradually merged, constituting a short sequence to ease the compute burden of self-attention at late stages.
	 }
	\vspace{-0mm}
	\label{fig:illustration_esvit}
\end{figure*}

\begin{table}[t]
\footnotesize
    \centering
    \resizebox{0.99\linewidth}{!}{
    \begin{tabular}{l|c|c|c|c}
    \toprule
            &  Stage 1 & Stage 2 & Stage 3 & Stage 4 \\
            \midrule
 Merging Rate  & $4\times$ & $8\times$ & $16\times$  &  $32\times$  \\     
 Feature Map  & $56\times56$ & $28\times28$ & $14\times14$  &  $7\times7$  \\    
 \midrule
    \multirow{3}{*}{Swin-T, $W\!=\!7$}  & concat $4\times4$ , $96$-d, LN    & concat $2\times2$, $192$-d, LN  & concat $2\times2$, $384$-d, LN  & concat $2\times2$, $ 768$-d, LN \\
    & \multicolumn{1}{c|}{$\left[\begin{array}{c} \text{window size}:7\times7  \\ \! 96\text{-d}~(3~\text{heads}) \! \end{array}\right]\! \times \! 2$} 
    & \multicolumn{1}{c|}{$\left[\begin{array}{c} \text{window size}:7\times7  \\ \! 192\text{-d}~(6~\text{heads}) \! \end{array}\right]\! \times \! 2$}
    & \multicolumn{1}{c|}{$\left[\begin{array}{c} \text{window size}:7\times7  \\ \! 384\text{-d}~(12~\text{heads}) \! \end{array}\right]\! \times \! 6$}
    & \multicolumn{1}{c}{$\left[\begin{array}{c} \text{window size}:7\times7  \\ \! 768\text{-d}~(24~\text{heads}) \! \end{array}\right]\! \times \! 2$} \\
\cmidrule{2-5}
    \multirow{3}{*}{Swin-S, $W\!=\!7$}  & concat $4\times4$ , $96$-d, LN    & concat $2\times2$, $192$-d, LN  & concat $2\times2$, $384$-d, LN  & concat $2\times2$, $ 768$-d, LN \\
    & \multicolumn{1}{c|}{$\left[\begin{array}{c} \text{window size}:7\times7  \\ \! 96\text{-d}~(3~\text{heads}) \! \end{array}\right]\! \times \! 2$} 
    & \multicolumn{1}{c|}{$\left[\begin{array}{c} \text{window size}:7\times7  \\ \! 192\text{-d}~(6~\text{heads}) \! \end{array}\right]\! \times \! 2$}
    & \multicolumn{1}{c|}{$\left[\begin{array}{c} \text{window size}:7\times7  \\ \! 384\text{-d}~(12~\text{heads}) \! \end{array}\right]\! \times \! 18$}
    & \multicolumn{1}{c}{$\left[\begin{array}{c} \text{window size}:7\times7  \\ \! 768\text{-d}~(24~\text{heads}) \! \end{array}\right]\! \times \! 2$} \\
\cmidrule{2-5}
    \multirow{3}{*}{Swin-B, $W\!=\!7$}  & concat $4\times4$ , $128$-d, LN    & concat $2\times2$, $256$-d, LN  & concat $2\times2$, $512$-d, LN  & concat $2\times2$, $ 1024$-d, LN \\
    & \multicolumn{1}{c|}{$\left[\begin{array}{c} \text{window size}:7\times7  \\ \! 128\text{-d}~(4~\text{heads}) \! \end{array}\right]\! \times \! 2$} 
    & \multicolumn{1}{c|}{$\left[\begin{array}{c} \text{window size}:7\times7  \\ \! 256\text{-d}~(8~\text{heads}) \! \end{array}\right]\! \times \! 2$}
    & \multicolumn{1}{c|}{$\left[\begin{array}{c} \text{window size}:7\times7  \\ \! 512\text{-d}~(16~\text{heads}) \! \end{array}\right]\! \times \! 18$}
    & \multicolumn{1}{c}{$\left[\begin{array}{c} \text{window size}:7\times7  \\ \! 1024\text{-d}~(32~\text{heads}) \! \end{array}\right]\! \times \! 2$} \\
 \cmidrule{1-5}
    \multirow{3}{*}{Swin-T, $W\!=\!14$}  & concat $4\times4$ , $96$-d, LN    & concat $2\times2$, $192$-d, LN  & concat $2\times2$, $384$-d, LN  & concat $2\times2$, $ 768$-d, LN \\
    & \multicolumn{1}{c|}{$\left[\begin{array}{c} \text{window size}:14\times14  \\ \! 96\text{-d}~(3~\text{heads}) \! \end{array}\right]\! \times \! 2$} 
    & \multicolumn{1}{c|}{$\left[\begin{array}{c} \text{window size}:14\times14  \\ \! 192\text{-d}~(6~\text{heads}) \! \end{array}\right]\! \times \! 2$}
    & \multicolumn{1}{c|}{$\left[\begin{array}{c} \text{window size}:14\times14  \\ \! 384\text{-d}~(12~\text{heads}) \! \end{array}\right]\! \times \! 6$}
    & \multicolumn{1}{c}{$\left[\begin{array}{c} \text{window size}:7\times7  \\ \! 768\text{-d}~(24~\text{heads}) \! \end{array}\right]\! \times \! 2$} \\
\cmidrule{2-5}
    \multirow{3}{*}{Swin-S, $W\!=\!14$}  & concat $4\times4$ , $96$-d, LN    & concat $2\times2$, $192$-d, LN  & concat $2\times2$, $384$-d, LN  & concat $2\times2$, $ 768$-d, LN \\
    & \multicolumn{1}{c|}{$\left[\begin{array}{c} \text{window size}:14\times14  \\ \! 96\text{-d}~(3~\text{heads}) \! \end{array}\right]\! \times \! 2$} 
    & \multicolumn{1}{c|}{$\left[\begin{array}{c} \text{window size}:14\times14  \\ \! 192\text{-d}~(6~\text{heads}) \! \end{array}\right]\! \times \! 2$}
    & \multicolumn{1}{c|}{$\left[\begin{array}{c} \text{window size}:14\times14  \\ \! 384\text{-d}~(12~\text{heads}) \! \end{array}\right]\! \times \! 18$}
    & \multicolumn{1}{c}{$\left[\begin{array}{c} \text{window size}:7\times7  \\ \! 768\text{-d}~(24~\text{heads}) \! \end{array}\right]\! \times \! 2$} \\
 \cmidrule{1-5}
     \multirow{3}{*}{ViL-T, $W\!=\!7$}  & concat $4\times4$ , $96$-d, LN    & concat $2\times2$, $192$-d, LN  & concat $2\times2$, $384$-d, LN  & concat $2\times2$, $ 768$-d, LN \\
    & \multicolumn{1}{c|}{$\left[\begin{array}{c} \text{window size}:7\times7  \\ \! 96\text{-d}~(3~\text{heads}) \! \end{array}\right]\! \times \! 2$} 
    & \multicolumn{1}{c|}{$\left[\begin{array}{c} \text{window size}:7\times7  \\ \! 192\text{-d}~(3~\text{heads}) \! \end{array}\right]\! \times \! 2$}
    & \multicolumn{1}{c|}{$\left[\begin{array}{c} \text{window size}:7\times7  \\ \! 384\text{-d}~(6~\text{heads}) \! \end{array}\right]\! \times \! 6$}
    & \multicolumn{1}{c}{$\left[\begin{array}{c} \text{window size}:7\times7  \\ \! 768\text{-d}~(12~\text{heads}) \! \end{array}\right]\! \times \! 2$} \\
\cmidrule{2-5}
    \multirow{3}{*}{CvT-T, $W\!=\!7$}  & concat $4\times4$ , $64$-d, LN    & concat $2\times2$, $192$-d, LN  & concat $2\times2$, $384$-d, LN  & concat $2\times2$, $ 768$-d, LN \\
    & \multicolumn{1}{c|}{$\left[\begin{array}{c} \text{window size}:7\times7  \\ \! 64\text{-d}~(1~\text{head}) \! \end{array}\right]\! \times \! 2$} 
    & \multicolumn{1}{c|}{$\left[\begin{array}{c} \text{window size}:7\times7  \\ \! 192\text{-d}~(3~\text{heads}) \! \end{array}\right]\! \times \! 2$}
    & \multicolumn{1}{c|}{$\left[\begin{array}{c} \text{window size}:7\times7  \\ \! 384\text{-d}~(6~\text{heads}) \! \end{array}\right]\! \times \! 6$}
    & \multicolumn{1}{c}{$\left[\begin{array}{c} \text{window size}:7\times7  \\ \! 768\text{-d}~(12~\text{heads}) \! \end{array}\right]\! \times \! 2$} \\    
            \bottomrule
    \end{tabular}}
     \vspace{2mm}
    \caption{Model configurations considered in our experiments. }
    \label{tab:model_config}
\end{table}


\section{Related Work}
\label{sec:related_work_appendix}

\paragraph{Self-supervised ConvNets.}
ConvNets-based SSL has been extensively studied in the literature. Based on the pre-training tasks, they can be broadly categorized into three classes:  Handcrafted pretext tasks~\citep{doersch2015unsupervised,noroozi2016unsupervised,pathak2016context,gidaris2018unsupervised,zhang2016colorful,larsson2016learning,zhang2017split,pu2016variational,donahue2019large}, contrastive learning~\citep{dosovitskiy2015discriminative,zhuang2019local,oord2018representation,hjelm2018learning,bachman2019learning,he2020momentum,chen2020simple,grill2020bootstrap} and prototype learning~\citep{caron2018deepcluster,caron2020unsupervised,li2020prototypical,xie2016unsupervised,yang2016joint,ji2019invariant,zhan2020online}. 
It is also known that data augmentations play a crucial role in SSL pipeline~\citep{chen2020improved,caron2020unsupervised,tian2020makes,li2020self}. 
The impact of pre-training dataset size/quality is explored for ConvNets in SSL~\citep{goyal2021self,tian2021dnc}.
To date, the search of best pre-taining tasks/datasets and augmentations are based on CNNs. Among them, SimCLR-v2~\citep{chen2020big}, BYOL~\citep{grill2020bootstrap} and SwAV~\citep{caron2020unsupervised} achieve the highest ImageNet linear probe performance with large ConvNet architectures. The performance tends to saturate with an increasingly growing model size, raising a question if ConvNets reach a limit in SSL.

\paragraph{Transformers for vision.} Vision Transformers (ViT)~\citep{dosovitskiy2020image} shows the great potentials of generalizing Transformers for computer vision, by achieving compelling accuracy in supervised learning, especially with large-scale data and high capacity models. DeiT~\citep{touvron2020training} further provides an effective ViT training strategy to ease the adaption of Transformers for practitioners.
Transformers have also been applied to other vision tasks, ranging from low-level tasks such as image generation~\citep{parmar2018image,chen2020generative} and enhancement~\citep{chen2020pre,yang2020learning}, to high-level tasks such as  object detection~\citep{carion2020end,zhu2020deformable,zheng2020end,dai2020up} and segmentation~\citep{wang2020max,wang2020end}, and to vision-language tasks~\citep{lu2019vilbert,tan2019lxmert,chen2019uniter,su2019vl,li2019visualbert,li2019unicoder,zhou2019unified,li2020oscar}.
Marrying Transformers with multi-stage architectures~\citep{vaswani2021scaling,wang2021pyramid,liu2021Swin,zhang2021vil,wu2021cvt} show higher classification accuracy in supervised learning, and enables applicability of Transformers for a broader range of vision tasks.
Given these properties, we believe multi-stage ViT is a must-study baseline for SSL in computer vision.


\paragraph{Discussion with other region-level tasks.} In Table~\ref{table:region_level_tasks}, we compare $\Lcal_R$ against the existing region-level tasks, including PIRL~\citep{misra2020self}, DenseCL~\citep{wang2020dense}, DetCo~\citep{xie2021detco}, InstLoc~\citep{yang2021instance}, PixPro~\citep{xie2021propagate}. Most of these region-level tasks improve object detection tasks, but hurt the ImageNet classification accuracy. DetCo achieves the best trade-off: improving the performance of both tasks, but with a sophisticated multi-level, global-local interaction algorithm. With the same number of pre-training epochs and augmented views, \shortname{} achieve the best ImageNet linear probe accuracy among all region-level tasks, with as minimum computational overhead as possible. This well serves our goal of building efficient SSL SoTA image classification system.

\begin{table}[t!]
  \centering
\scriptsize
  \scalebox{1.0}{
  \begin{tabular}{l | p{1.0cm} | p{2.8cm} | p{2.4cm} | p{2.8cm} | p{1.0cm}}
    \toprule
Name & Framework & Pre-train Task Description & 	 Major Motivation & Major Downstream Task Performance & ImageNet    
    \\
    \midrule
PIRL 
& Contrastive
& Jigsaw pretext task in a way that encourages the image representations to be invariant to the image patch perturbation	
& General-purpose visual backbone learning	Mostly for image classification. 
& Image classification
&  63.6\% \\
    \midrule
DenseCL	
& Contrastive 
& A pairwise contrastive (dis)similarity loss at the patch level between two views. A queue of negative sample features from other images is maintained	
& Improving dense visual prediction task performance
& Improving object detection performance.
& 63.6\% \\
    \midrule
DetCo	
& Contrastive
& Multi-level features with three contrastive tasks between global images and local patches are considered: global-global, global-local, local-local.	
& Improving dense visual prediction task performance	
& Improving object detection performance. DetCo achieves the best performance trade-off on both classification and detection. 
& 68.6\% \\
    \midrule
PixPro
& Contrastive
& Features from the two views are encouraged to be consistent between a regular patch representation and a smoothed patch representation within the same image.
& Improving dense visual prediction task performance	
& Mostly focusing on the improved performance on detection tasks. 
& 66.3\% \\
    \midrule
InstLoc	
& Contrastive 
& image instances are pasted at various locations and scales onto background images. The pretext task is to predict the instance category given the composited images and the foreground bounding boxes	
& Improving dense visual prediction task performance	
& Mostly focusing on the improved performance on detection tasks. 
& 61.7\%  \\
\midrule
EsViT (ours) 
& Non-contrastive
& A pairwise cross-entropy loss at the patch level between two positive views. No need/interaction with other images in the batch (eg, no negative samples)	
& Recovering the automatic correspondence learning property of self-supervised monolithic transformers, and thus improving learning efficiency.
& Consistently improving image classification tasks.  It creates new SoTA 81.3\% on ImageNet linear probe accuracy, showing 3.5x parameter-efficient and has 10x higher throughput than previous SoTA MoCo-v3. Reporting 75.7\% ImageNet linear probe performance for ResNet-50.
& {\bf 69.9\%} \\
    \bottomrule
  \end{tabular}
  }
  \vspace{1mm}
\caption{Discussion of related works on various region-level tasks. The last columns reports the ImageNet linear probe performance for ResNet-50 trained with 2 augmented views for 200 epochs.}
  \label{table:region_level_tasks}  
  \vspace{-2mm}
\end{table}

\section{Experiments}
\subsection{Throughput estimate and conversion}

Since different papers report throughput on different hardwares, it is not ready to compare the numbers directly. Noting that all papers report the throughput for ViT-B/DeiT-B, we use this number to align and convert the throughput. In Table~\ref{table:throughput_conversion}, we describe our process and results of standardizing the throughput.

\begin{table}[t!]
  \centering
  \scalebox{0.90}{
  \begin{tabular}{l | p{2.8cm} | p{3.8cm} | p{2.8cm}  | p{1.5cm}}
    \toprule
    Data source     &   Table 2 in DINO~\citep{caron2021emerging}  &  Table 3 in  MLP-Mixer~\citep{tolstikhin2021mlp}  &  Table 1 in Swin~\citep{liu2021Swin} & Our runs \\
    \midrule
  DeiT-S~/~$P\!=\!16$ & \textcolor{blue}{1007}   &   & \textcolor{blue}{940.4}    & \\
  DeiT-B~/~$P\!=\!16$ & \textcolor{blue}{312}    &    & \textcolor{blue}{292.3}   &   \\
  DeiT-S~/~$P\!=\!8$   &\textcolor{blue}{180}   &   &    &   \\
  DeiT-B~/~$P\!=\!8$   & \textcolor{blue}{63}   &    &    &   \\  
  ViT-B~/~$P\!=\!16$   & \textcolor{blue}{312}    & \textcolor{blue}{861}  &  & \\
  ViT-S~/~$P\!=\!16$   & \textcolor{orange}{102}  &  \textcolor{blue}{280}  &  & \\
  ViT-H~/~$P\!=\!14$   & \textcolor{orange}{32}  & \textcolor{blue}{87} & & \\
  ViT-L~/~$P\!=\!7$   & \textcolor{orange}{17}  & \textcolor{orange}{47}$^{\dagger}$ & & \\   
  Swin-T~/~$W\!=\!7$    & \textcolor{orange}{808}  &  & \textcolor{blue}{755.2} & \textcolor{green!50!black}{726.13} \\  
  Swin-S~/~$W\!=\!7$    & \textcolor{orange}{467}  &  & \textcolor{blue}{436.9}  &  \\ 
  Swin-B~/~$W\!=\!7$    & \textcolor{orange}{297}  &  & \textcolor{blue}{278.1}  &  \\    
  Swin-T~/~$W\!=\!14$    & \textcolor{orange}{660}  &  &   & \textcolor{green!50!black}{593.24}  \\  
  Swin-S~/~$W\!=\!14$    & \textcolor{orange}{383}  &  &   & \textcolor{green!50!black}{344.20}  \\ 
  Swin-B~/~$W\!=\!14$    & \textcolor{orange}{254}  &  &    & \textcolor{green!50!black}{228.36}  \\      
  ViL-T~/~$W\!=\!7$    & \textcolor{orange}{386}  &  &    & \textcolor{green!50!black}{346.72}   \\     
  CvT-T~/~$W\!=\!7$    & \textcolor{orange}{848}  &  &    & \textcolor{green!50!black}{761.89}  \\     
    \bottomrule
  \end{tabular}
  }
  \vspace{1mm}
\caption{Throughput estimate and standardization. All \textcolor{orange}{numbers in orange} are estimated/converted, while \textcolor{blue}{numbers in blue} are collected from the papers, and  \textcolor{green!50!black}{numbers in green} are runs on our machines.
All papers report the throughput of ViT-B or DeiT-B, which are essentially the same model. We use this fact to align the throughput reported in different papers. $^{\dagger}$ This number is estimated via the statement in~\citep{chen2021empirical} that ``reducing the patch size to $7\times7$ keeps the model size unchanged, but increases FLOPs to $\sim 6\times$''. All numbers are standardized into throughput reported by~\citep{caron2021emerging}. }
  \label{table:throughput_conversion}  
  \vspace{-2mm}
\end{table}

\subsection{The computation and memory overhead of the proposed $\Lcal_R$}
\label{sec:lr_efficiency}
We emphasize that adding $\Lcal_{R}$ to the multi-stage transformer architectures yields acceptable extra computational cost, while adding $\Lcal_{R}$  directly to the monolithic transformer architectures has a huge computational overhead. To demonstrate this, we report the cost comparisons in Table~\ref{tab:efficiency_lr}. For each setting, we report [Memory Usage (MB) / Running time per iteration (second/iteration)]. In Table Table~\ref{tab:efficiency_lr} (a), when the batch size is gradually increased (\eg, batch-size=12), the memory cost increases nearly 4 times for monolithic architectures, while increases 1.6 times for multi-stage architectures. Similar trends are shown for training cost per iteration increase ratio (1.47 vs 1.15). This indicates $\Lcal_{R}$ can more naturally fit multi-stage architectures.

Similarly, we compare computational cost comparisons [Memory Usage (MB) / Running time per iteration (second/iteration)] in Table~\ref{tab:efficiency_lr} (b), for other network architecture configurations . From the increased cost ratio, we see that $\Lcal_{R}$ adds  acceptable cost in terms of both memory and training time, compared with the baseline.

\begin{table}[t!]
\begin{subtable}{1.0\linewidth}
\centering
\scalebox{0.87}{
\begin{tabular}{ c  @{\hspace{6pt}}c@{\hspace{12pt}}|l@{\hspace{12pt}}l@{\hspace{12pt}}l@{\hspace{12pt}}l@{\hspace{12pt}}c}
\toprule
&	& &	Batch-Size=1 &	Batch-Size=8 &	Batch-Size=12 \\
\midrule
\multirow{2}{*}{Monolithic} & 
 ViT-S  &  $\Lcal_{V}$ 	&	1391 / 0.209036	& 2735 / 0.234993 &	3533 / 0.238160 \\
&  ViT-S 	& $\Lcal_{V}\!+\!\Lcal_{R}$ &	2641 / 0.233150	& 10358 / 0.321155 &	14128 / 0.352339 \\
& {\bf Increased Cost Ratio} & &	{\bf 1.8986 / 1.1153} & {\bf 3.7872 / 1.3666} &	{\bf 3.9988 / 1.4794} \\
\midrule
\multirow{2}{*}{Multi-stage} & 
 Swin-T  &  $\Lcal_{V}$ &	1704 / 0.285451	& 4229 / 0.323884 &	5611 / 0.366367 \\
&  Swin-T  & $\Lcal_{V}\!+\!\Lcal_{R}$ &	2346 / 0.301672	& 6232 / 0.374634 &	8917 / 0.421135 \\
& {\bf Increased Cost Ratio} & &	{\bf 1.3767 / 1.0568}	& {\bf 1.4736 / 1.2323} &	{\bf 1.5889 / 1.1494} \\
\bottomrule
\end{tabular} 
}
\caption{Comparisons for increased batch sizes. }  \vspace{2mm}
\end{subtable}
\begin{subtable}{0.99\linewidth}
\centering
\scalebox{0.87}{
\begin{tabular}{ @{\hspace{-0pt}}l@{\hspace{12pt}}|c@{\hspace{12pt}}c@{\hspace{12pt}}c@{\hspace{12pt}}c@{\hspace{12pt}}c}
\toprule
EsViT	& $\Lcal_{V}$ &	$\Lcal_{V}\!+\!\Lcal_{R}$ &	{\bf Increased Cost Ratio} \\ 
\midrule
Tiny (W=7) & 1704 / 0.285451 & 	2346 / 0.301672	& {\bf 1.3767 / 1.0568}\\ 
Small (W=7) & 	2685 / 0.501876 & 	3132 / 0.535203 & 	{\bf 1.1664 / 1.0664}\\ 
Base (W=7)& 	3726 / 0.516058	& 4374 / 0.550617 & 	{\bf 1.1739 / 1.0669}\\ 
Tiny (W=14)& 	2159 / 0.288118	& 2801 / 0.310108 & 	{\bf 1.0890 / 1.0763}\\ 
Small (W=14)& 	3518 / 0.496823	& 4153 / 0.521739 & 	{\bf 1.1805 / 1.0501}\\ 
Base (W=14)& 	5032 / 0.511701	& 5681 / 0.537826 & 	{\bf 1.1289 / 1.0510}\\ 
\bottomrule
\end{tabular}
 }
\caption{Comparisons for various network architecture configurations. 
}
\end{subtable}
\vspace{2mm}
    \caption{ Computational cost comparisons in the format of [{\em Memory Usage (MB) / Running time per iteration (second/iteration)}].}
    \label{tab:efficiency_lr}

\end{table}

\subsection{Experimental settings of pre-training and evaluation on ImageNet}
\label{sec:experimental_settings}
We study unsupervised pre-training performed in ImageNet-1K dataset~\citep{deng2009imagenet} without labels. The default training details are described as follows, mostly following~\citep{caron2021emerging}. We train with the Adamw optimizer~\citep{loshchilov2018fixing}, a batch size of $512$, and total epochs $300$. Linear warmup of the learning rate is used during the first 10 epochs, with its base value determined with the linear scaling rule~\citep{goyal2017accurate}: $lr = 0.0005*\text{batchsize}/256$. After this warmup, the learning rate is decayed with a cosine schedule. 
We build our systems based on Swin Transformers~\citep{liu2021Swin} in our experiments.  
Swin-B has a model size and computation complexity similar to ViT-B/DeiT-B (patch size 16). We also considered Swin-T and Swin-S, which have the complexity that are similar to those of ResNet-50 (DeiT-S) and ResNet-101, respectively. The default window size is $W\!=\!7$.

One major common protocol to evaluate SSL is linear probe on ImageNet-1K, where features are extracted from a frozen backbone, and a supervised linear classifier is trained. For all Transformer models, we use the concatenation of view-level features $\bar{z}$ in the last 4 layers (the results are similar to the use of last 3 or 5 layers in our initial experiments).


\begin{table}[t!]
  \centering
  \scalebox{0.90}{
\begin{tabular}{c | c c c c c } 
 \toprule
 Dataset & Classes & Train size & Test size & Evaluation metric & Source link \\ [0.5ex] 
 \midrule
 Food-101 & 102 & 75,750 & 25,250 & Accuracy & \href{https://www.tensorflow.org/datasets/catalog/food101}{Tensorflow} \\ 
 CIFAR-10 & 10 & 50,000 & 10,000 & Accuracy & \href{https://www.tensorflow.org/datasets/catalog/cifar10}{TensorFlow} \\
 CIFAR-100 & 100 & 50,000 & 10,000 & Accuracy & \href{https://www.tensorflow.org/datasets/catalog/cifar100}{TensorFlow} \\
SUN397 & 397 & 19,850 & 19,850 & Accuracy & \href{https://www.tensorflow.org/datasets/catalog/sun397}{Tensorflow} \\
Stanford Cars & 196 & 8,144 & 8,041 & Accuracy & \href{https://ai.stanford.edu/~jkrause/cars/car_dataset.html}{Stanfold Cars} \\
FGVC Aircraft (variants) & 100 & 6,667 & 3,333 & Mean-per-class & \href{https://www.robots.ox.ac.uk/~vgg/data/fgvc-aircraft/}{FGVC website} \\
VOC2007 classification & 20 & 5,011 & 4,952& 11-point mAP & \href{http://host.robots.ox.ac.uk/pascal/VOC/voc2007/index.html}{voc2007} \\
Describable Textures & 47& 3,760& 1,880& Accuracy& \href{https://www.tensorflow.org/datasets/catalog/dtd}{TensorFlow} \\
Oxford-IIIT  Pets & 37& 3,680& 3,669& Mean-per-class& \href{https://www.robots.ox.ac.uk/~vgg/data/pets/}{Oxford-IIIT Pet} \\
Caltech-101& 102& 3,060& 6084& Mean-per-class& \href{https://www.tensorflow.org/datasets/catalog/caltech101}{TensorFlow} \\
Oxford Flowers 102& 102& 2,040& 6,149 & Mean-per-class& \href{https://www.tensorflow.org/datasets/catalog/oxford_flowers102}{TensorFlow} \\
MNIST& 10& 60,000& 10,000 & Accuracy& \href{https://www.tensorflow.org/datasets/catalog/mnist}{TensorFlow} \\
 Facial Emotion Recog. 2013 $^{\ast}$& 8& 32,298& 3,589 & Accuracy& \href{https://www.kaggle.com/c/challenges-in-representation-learning-facial-expression-recognition-challenge/data}{Kaggle fer2013}\\
STL10 & 10& 5,000& 8,000& Accuracy& \href{https://www.tensorflow.org/datasets/catalog/stl10}{TensorFlow} \\
GTSRB $^{\ast}$& 43 & 26,728 & 12,630 & Accuracy & \href{https://benchmark.ini.rub.de/gtsrb_dataset.html}{GTSRB website} \\
PatchCamelyon & 2 & 294,912 & 32,768 & Accuracy & \href{https://www.tensorflow.org/datasets/catalog/patch_camelyon}{TensorFlow} \\
UCF101 $^{\ast}$ & 101 & 9,537 & 3783 & Accuracy & \href{https://www.tensorflow.org/datasets/catalog/ucf101}{TensorFlow} \\
Hateful Memes & 2 & 8,500 & 500 & ROC-AUC & \href{https://ai.facebook.com/blog/hateful-memes-challenge-and-data-set/}{FaceBook} \\
\bottomrule
\end{tabular}
}
\vspace{2mm}
\caption{A suite of 18 datasets used in linear probe.$^{\ast}$ indicates dataset whose train/test size we obtained is slightly different from Table 9 in~\citep{radford2021learning}. }
\label{table:lp_dataset}
\end{table}
\begin{table}[t!]
  \centering
  \scalebox{0.90}{
\begin{tabular}{c | c c c c c } 
 \toprule
\multirow{2}{*}{Methods}   & CLIP & Supervised & Supervised$^{\ddagger}$ &  Supervised & \shortname{} \\ [0.5ex] 
  & ResNet-50 & ResNet-50 & ResNet-50 &  Swin-T & Swin-T \\ 
 \midrule
 Food-101 &  86.4 & 71.3  & 71.3 &  77.4 &  80.0 \\ 
 CIFAR-10 & 88.7   & 91.8  &   91.8  & 94.0  & 95.3 \\
 CIFAR-100 & 70.3 &  74.5 &   74.5  &  77.5 &  82.2\\
SUN397 & 73.3 &  60.5 &  60.3   &  64.3 & 67.6 \\
Stanford Cars & 78.3 & 49.9 &  50.1  & 55.3  & 66.4 \\
FGVC Aircraft (variants) & 49.1 &  48.5 &  48.4 &  51.5 & 61.1 \\
VOC2007 classification & 87.1 & 83.8  & 83.6  & 84.2  &  85.5  \\
Describable Textures & 76.4 & 72.3 &  72.6 & 73.1  & 78.1  \\
Oxford-IIIT  Pets & 88.2 & 92.4   &  92.1   & 93.3  &  92.8  \\
Caltech-101 & 89.6 &  90.8 & 90.4  &  90.8  &  93.0  \\
Oxford Flowers 102   &  96.1 & 90.8  &  91.1 & 91.5  & 97.4 \\
MNIST & 98.3  & 98.3  &  98.3 &  98.3 &   98.3  \\
 Facial Emotion Recog. 2013  &  64.2 & 54.9 &  55.9 & 55.1  & 59.3  \\
STL10  & 97.2  & 96.4 &  97.0 & 97.9  & 98.9  \\
GTSRB  & 82.4 & 70.6 & 75.7 & 72.9  & 84.3  \\
PatchCamelyon  & 82.7  & 82.5  & 82.6 & 84.0  & 84.6   \\
UCF101  & 81.6 & 71.2 & 72.1  &  79.0 &  81.1 \\
Hateful Memes  & 65.7 & 56.5 & 49.9 & 51.2  & 52.0  \\
\midrule
Average  & 80.86 & 75.39 & 75.43 &  77.29 &  80.99 \\
\bottomrule
\end{tabular}
}
\vspace{2mm}
\caption{The  linear probe results on 18 datasets at the scale of ResNet-50/Swin-T. $^{\ddagger}$ indicates the results reproduced by us, which verifies that our implementation pipeline is consistent with~\citep{radford2021learning}.}
\label{table:lp_results_appendix}
\end{table}

\begin{table}[t!]
  \centering
  \scalebox{0.90}{
  \begin{tabular}{l | p{2.8cm} | p{2.8cm} | p{2.8cm} }
    \toprule
    Method   &  View-level &   Region-level &   Top-1 Accuracy (\%)  \\
    \midrule
\multicolumn{4}{l}{\em \hspace{0mm} Performance comparison of ResNet-50 with 200 epochs and 2 augmented views}  \\
  MoCo-v2  & Contrastive & - &   67.5  \\
  DenseCL  & Contrastive & Contrastive &   63.6  \\
  DetCo    & Contrastive & Contrastive &   68.6  \\ 
  DINO     & Non-Contrastive & - &   69.2  \\ 
\rowcolor{Gray}
  \shortname{}  & Non-Contrastive & Non-Contrastive &  {\bf 69.9} \\ 
  \midrule
\multicolumn{4}{l}{\em \hspace{0mm} SoTA performance comparison of ResNet-50 with numbers and settings reported in each paper}   \\
  MoCo-v2 (800 epochs) & Contrastive & - &   72.2 \\
SwAV (800 epochs, w/ multi-crop) & Contrastive & -  & 75.3  \\  
  Barlow Twins (1000 epochs) & - & - &   73.2 \\
  VICReg (1000 epochs) & - & - &   73.2 \\
SimSiam (800 epochs, 2 views) & Non-Contrastive & - & 71.3 \\
BYOL (1000 epochs, w/ multi-crop) & Non-Contrastive & -  & 74.3  \\
DINO (300 epochs, w/ multi-crop) & Non-Contrastive & -  & 75.0  \\
\rowcolor{Gray}
\shortname{} (300 epochs, w/ multi-crop) & Non-Contrastive & Non-Contrastive  & {\bf 75.7}  \\
    \bottomrule
  \end{tabular}
  }
  \vspace{1mm}
\caption{Linear probe performance of a ResNet-50 network with different SSL methods. }
  \label{table:r50_linear_probe}  
  \vspace{-2mm}
\end{table}

\subsection{Comparison with a ResNet-50 backbone}
To compare our \shortname{} learning method with other SSL algorithms, we conduct experiments with a ResNet-50 backbone, and show the results in Table~\ref{table:r50_linear_probe}.

\subsection{Linear probe on a suite of small datasets}
\paragraph{Datasets.} Table \ref{table:lp_dataset} shows details and source of all datasets used for linear probe, including the number of classes, the size of training set and testing set, metrics used in evaluation, as well as a public source of the dataset.
Note that original  UCF101 dataset is a video dataset. Here the middle frame of each video is extracted to form a classification dataset. There are 3 train/val splits in Tensorflow, we use the first one.

\paragraph{Automatic hyper-parameter tuning.} We rigorously follow~\citep{radford2021learning} to conduct training and evaluation for linear probe on the downstream datasets. 
We train a logistic regression classifier using scikit-learn's L-BFGS implementation, with maximum $1,000$ iterations, and report the corresponding metric for each dataset. We determine the $L_2$ regularization strength $\lambda$ using a hyperparameter sweep on the validation
sets over the range between $10^{-6}$ and $10^6$ , with 96 logarithmically spaced steps. To save compute required for the sweeps, we perform a parametric binary search that starts with 
$ \lambda = [10^{-6}, 10^{-4}, 10^{-2}, 1, 10^2, 10^4, 10^6] $ and iteratively halves the interval around the peak until it reaches a resolution of 8 steps per decade. The hyperparameter sweeps are performed on a validation split of each dataset. For the datasets that contain a validation split in addition to a test split, we use the provided validation set to perform the hyperparameter search, and for the datasets that do not provide a validation split or have not published labels for
the test data, we split the training dataset to perform the hyperparameter search. For the final result, we combine the validation split back with the training split and report the performance on the unused split.

\paragraph{Detailed results.} 
Only the last layer feature is considered for all models for simplicity, though adding features from more layers may potentially improve the results. Table~\ref{table:lp_results_appendix} shows the results for architectures at a similar scale of ResNet-50 or Swin-T. 
The first two columns are numbers from~\citep{radford2021learning}.
CLIP with ResNet-50 is pre-trained on  400 million image-text pairs. Supervised ResNet-50 and Swin-T are pre-trained on ImageNet-1K, on which \shortname{} with Swin-T is pre-trained as well (Batch Size=512). \shortname{} outperforms its supervised counterpart, and is on par with the performance of CLIP in a similar image encoder architecture scale.

\subsection{Pre-training datasets}

We describe the statistics and training schedule on larger and less curated datasets in Table~\ref{table:pretraining_datasets}.  The pre-training epochs are chosen so that the model is trained with a similar number of augmented views.

\begin{table}[h!]
  \centering
  \scalebox{0.70}{
  \begin{tabular}{lp{8cm} l | p{9mm} p{9mm}}
    \toprule
    Name     &   Description  &  Size (\#Images)  &  Epochs & Warmup \\
    \midrule
    ImageNet-1K~\citep{deng2009imagenet} & Images evenly distributed in 1K object concepts & $1.2$ million   &  300 & 10 \\
    WebVision-v1~\citep{li2017webvision}    & Web images with 1K concept queries from ImageNet-1K        & $2.4$ million & 150  & 5 \\
    OpenImages-v4~\citep{kuznetsova2020open}    & Diverse/complex scenes with several objects for detection  & $7.5$ million & 50 & 2 \\
    ImageNet-22K~\citep{deng2009imagenet}   & Images distributed in 22K object concepts in a hierarchy & $14.2$ million  & 30  & 1 \\    
    \bottomrule
  \end{tabular}
  }
  \vspace{1mm}
\caption{Pre-train dataset statistics and training schedule.}
  \label{table:pretraining_datasets}  
  \vspace{-2mm}
\end{table}

\subsection{Results on correspondence learning}
\label{sec:correspondence_imagenet}
\begin{figure*}[t!]
	\vspace{-0mm}\centering
	\begin{tabular}{c c}
		\hspace{-2mm}
		\includegraphics[height=3.2cm]{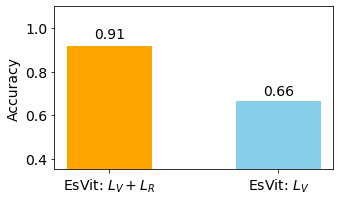} & 
		\hspace{3mm}
		\includegraphics[height=3.2cm]{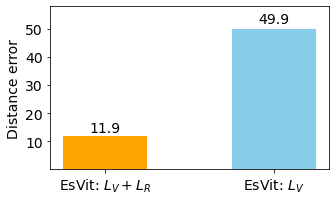} 
		\\			
		(a) Accuracy
		 & 
		(b) Distance error
	\end{tabular}
	\vspace{-0mm}
	\caption{Quantitative evaluation on correspondence learning on ImageNet validation set. $\Lcal_R$ can significantly improve correspondence learning quality for multi-stage architectures.  As a reference, DINO ($\Lcal_V$ with monolithic Transformer architecture) achieves 0.95 accuracy and 2.49 distance error, which we believe is a strong evidence to identify the intriguing property of automatic correspondence learning.}
	\vspace{-1mm}
	\label{fig:correspondence_metrics}
\end{figure*}

\begin{figure*}[t!]
	\vspace{-0mm}\centering
	\begin{tabular}{c c c}
		\hspace{-2mm}
		\includegraphics[height=2.3cm]{figs/correspondence/pairs_appendix/cute_dog_1_dino.png}  & 
		\hspace{-3mm}
		\includegraphics[height=2.2cm]{figs/correspondence/pairs_appendix/cute_dog_1_lv.png} & 
		\hspace{-3mm}
		\includegraphics[height=2.2cm]{figs/correspondence/pairs_appendix/cute_dog_1_lr.png} 		
		\\	
		\hspace{-1mm}
		\includegraphics[height=2.3cm]{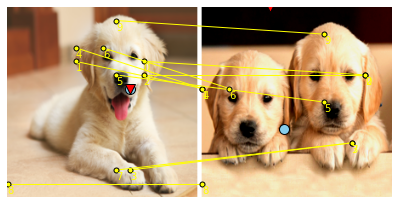}  & 
		\hspace{-3mm}
		\includegraphics[height=2.3cm]{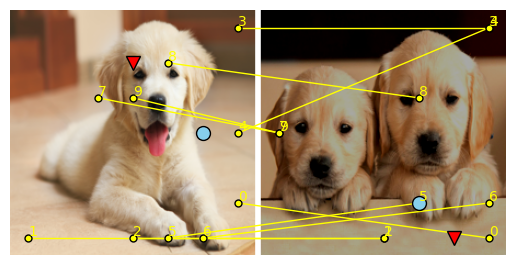} & 
		\hspace{-3mm}
		\includegraphics[height=2.3cm]{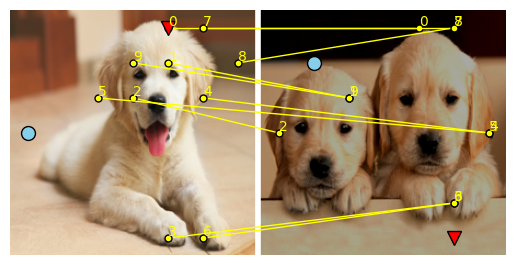} 		
		\\			
		\hspace{-2mm}
		\includegraphics[height=2.2cm]{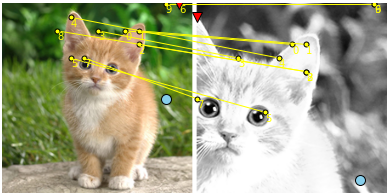}  & 
		\hspace{-3mm}
		\includegraphics[height=2.2cm]{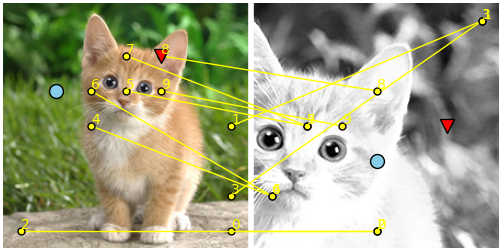} & 
		\hspace{-3mm}
		\includegraphics[height=2.2cm]{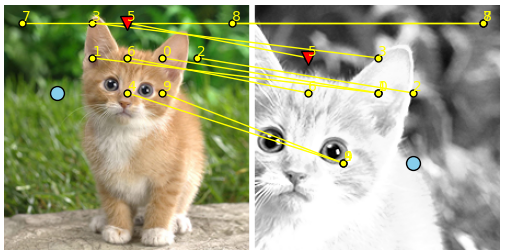} 		
		\\
		\includegraphics[height=2.2cm]{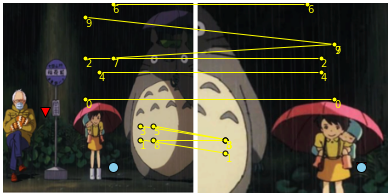}  & 
		\hspace{-3mm}
		\includegraphics[height=2.2cm]{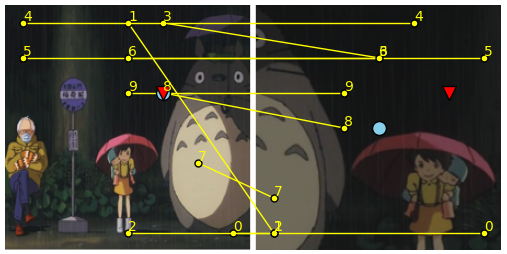} & 
		\hspace{-3mm}
		\includegraphics[height=2.2cm]{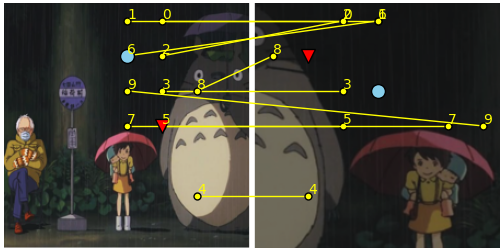} 		
		\\	
		\includegraphics[height=2.2cm]{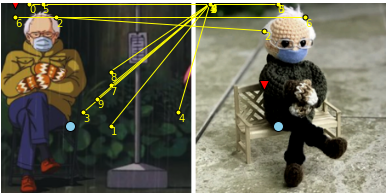}  & 
		\hspace{-3mm}
		\includegraphics[height=2.2cm]{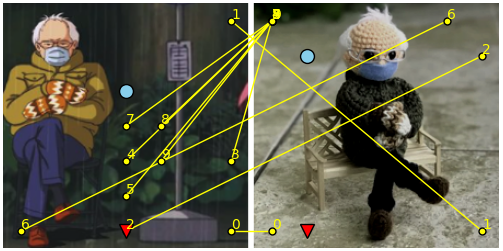} & 
		\hspace{-3mm}
		\includegraphics[height=2.2cm]{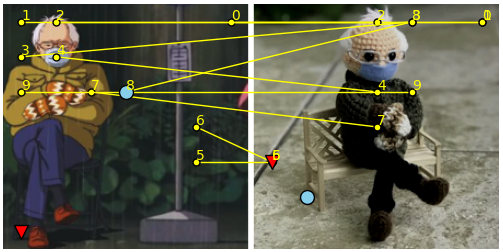} 		
		\\	
		\includegraphics[height=2.2cm]{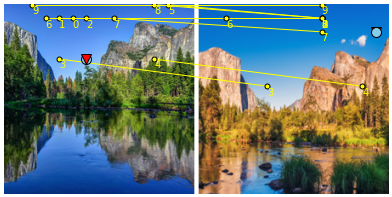}  & 
		\hspace{-3mm}
		\includegraphics[height=2.2cm]{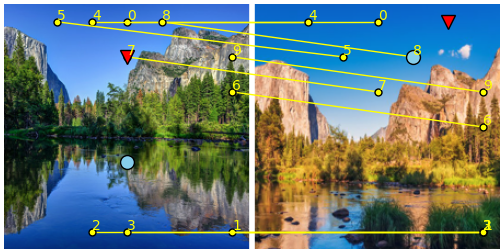} & 
		\hspace{-3mm}
		\includegraphics[height=2.2cm]{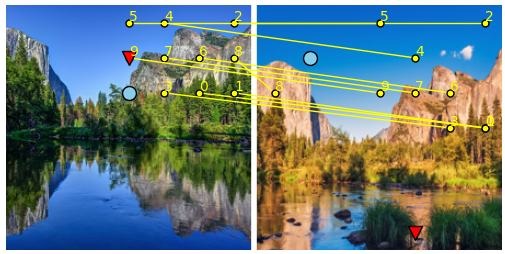} 		
		\\			
		(a) DINO: DeiT-S  & 
		\hspace{-2mm}
		(b) \shortname{}: $\Lcal_V$    \vspace{2mm} & 
		\hspace{-2mm}		
		(c) \shortname{}: $\Lcal_V\!+\!\Lcal_R$   \hspace{-0mm} \\ 
	\end{tabular}
	\vspace{-3mm}
	\caption{The learned correspondences.  {\bf \textcolor{yellow!80!black}{Yellow}}  lines are the top-10 correspondences between two views, where the numbers indicates the rankings of similarity scores, yellow dots with the same number are paired. The  {\bf \textcolor{blue!50}{blue}} dot and {\bf \textcolor{red}{red}}  triangle indicates the most similar local regions that correspond to the global feature of the view itself and the other view, respectively. Please zoom in for detailed correspondence mappings.
	 }
	\vspace{-1mm}
	\label{fig:correspondences_appendix}
\end{figure*}

We first quantitatively evaluate the correspondence learning results with 50K images in the ImageNet validation dataset. We create a simple evaluation dataset with mild augmentations. For a center-crop image, we apply \texttt{HorizontalFlip}, then \texttt{ColorJitter} and \texttt{RandomGrayscale} to create a new augmented view. In this way, ground-truth correspondences are created. Please see the 1st row of Figure~\ref{fig:correspondences_appendix} for one such example. The top-10 correspondences are used for evaluation. Two metrics are considered: (1) Accuracy measures the percentage of correctly matched region pairs, (2) distance error indicates the averaged $\ell_2$ distance between the predicted matched region and ground-truth region (the value is 0 for perfect matching). The results are reported in Figure~\ref{fig:correspondence_metrics}. DINO with monolithic Transformers shows surprisingly good performance on correspondence learning. The use of multi-stage Transformer architecture  reduces this ability, shows a lack of good region correspondence. With $\Lcal_R$, the region matching ability is significantly recovered.

In Figure~\ref{fig:correspondences_appendix}, we visualize the correspondences for more images. Overall, DINO with monolithic Transformers is able to discover most salient correspondences of semantic meaning in the mild augmentation conditions, even without an implicit region matching loss in training. We believe this previously underestimated property is whole-noting, and has potentials to enable more applications. However, this desired property gets dilated when changing from monolithic to multi-stage Transformer architecture (from column 1 to column 2), then the proposed region level task can alleviate this issue (from column 2 to column 3).  

To more specifically analyze the correspondences, we note the following results. The first row shows a simple case, where only images of left-to-right flipped views are presented. The ground-truth correspondences should be horizontal lines that link the two flipped regions. It reveals that the view-level pre-train task alone is insufficient to learn good correspondences for the multi-stage Transformer architecture, while region matching task can alleviate this issue significantly. Similar observations are shown in row 3 and row 4.

We further study more cases that requires real-world correspondences in row 2, row 5 and row 6. These views are not generated with data augmentation (as in model pre-training), but are often presented in more practical scenarios: one-to-many mappings, cartoon-to-toy,  seasonal changing of the scene, respectively. The proposed region matching task can work particularly well in those cases.


\subsection{More visualization results of attention maps}
\begin{figure*}[t!]
	\vspace{-0mm}\centering
	\begin{tabular}{c c c}
	\multicolumn{3}{c}{\includegraphics[height=2.6cm]{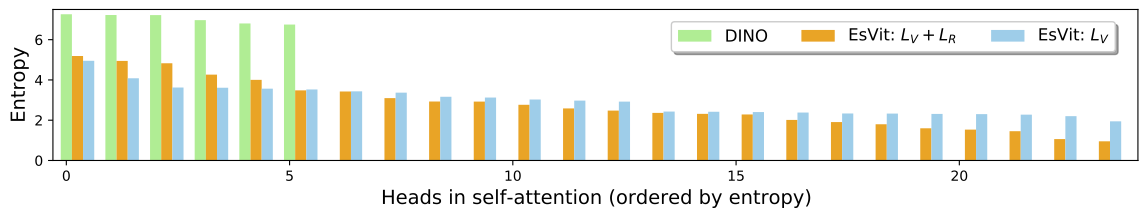}} 	\hspace{-4mm} \\
		\hspace{-2mm}
		\includegraphics[height=7.2cm]{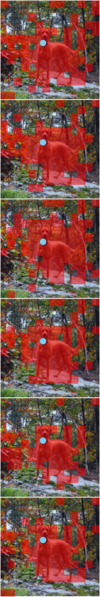}  & 
		\hspace{-3mm}
		\includegraphics[height=7.2cm]{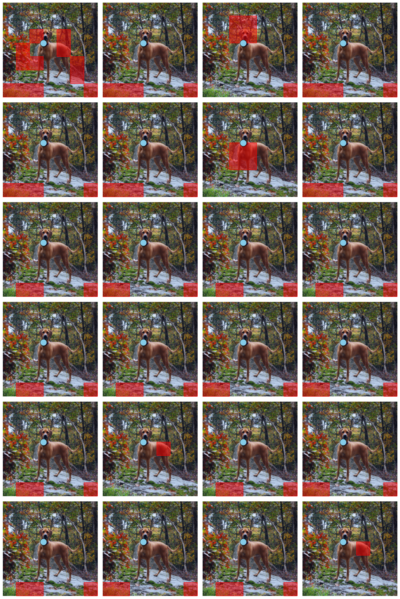} & 
		\hspace{3mm}
		\includegraphics[height=7.2cm]{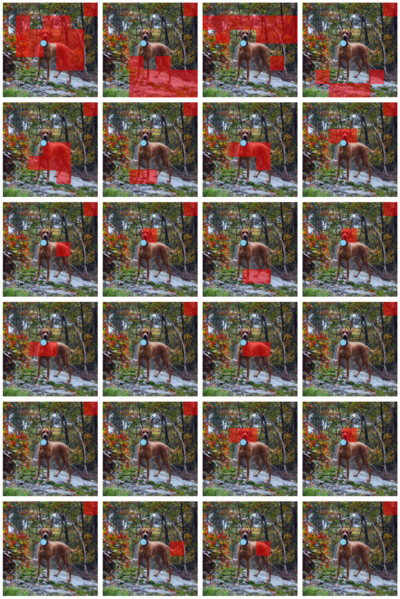} 		
		\\	
		\hspace{-2mm}
		\includegraphics[height=7.2cm]{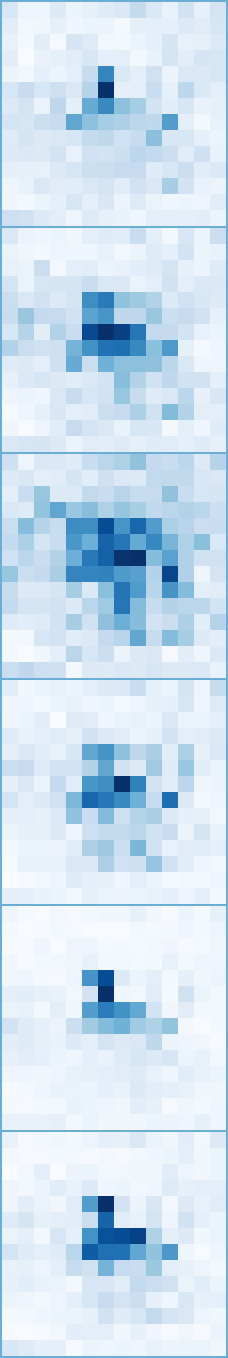}  & 
		\hspace{-2mm}
		\includegraphics[height=7.2cm]{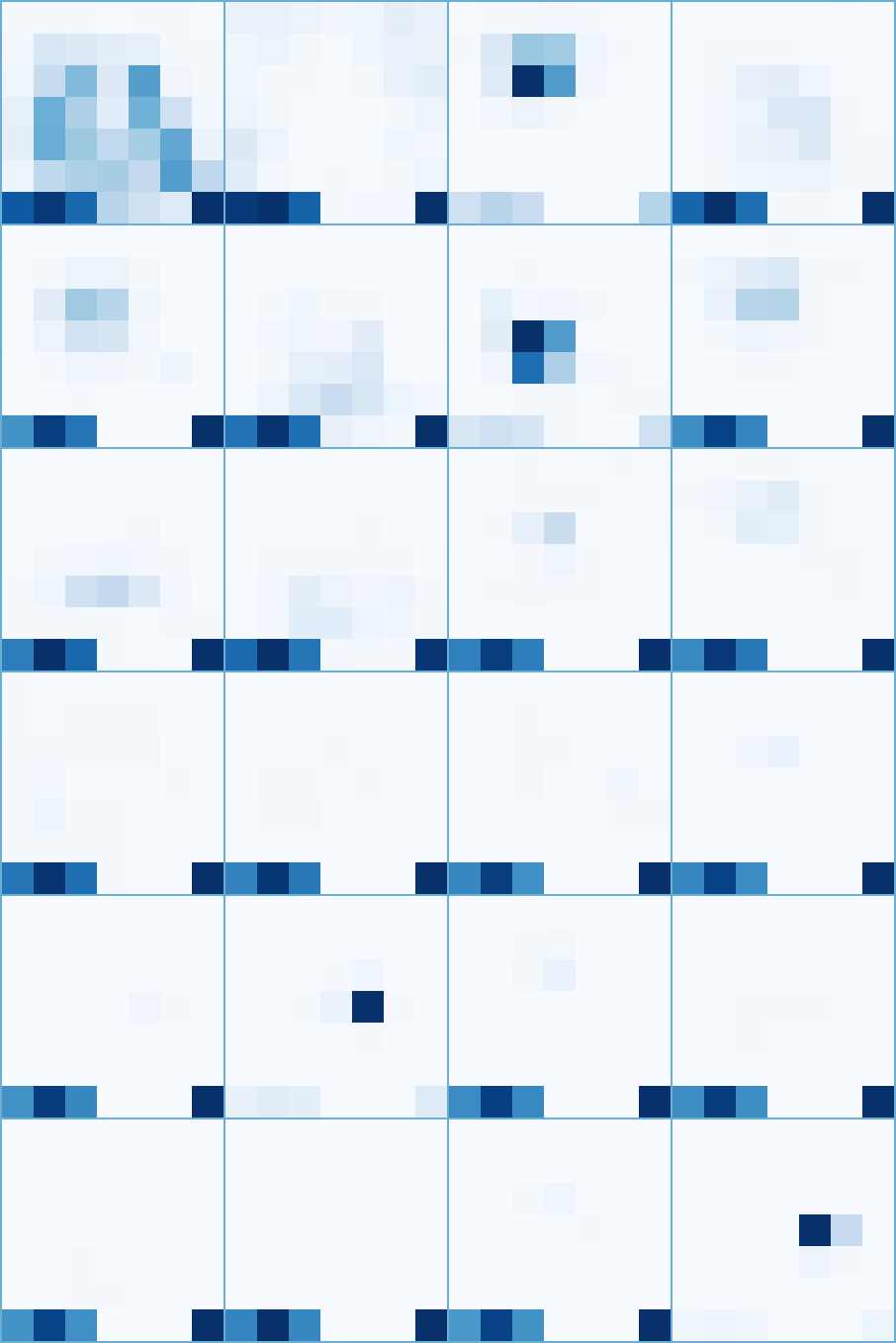} & 
		\hspace{3mm}
		\includegraphics[height=7.2cm]{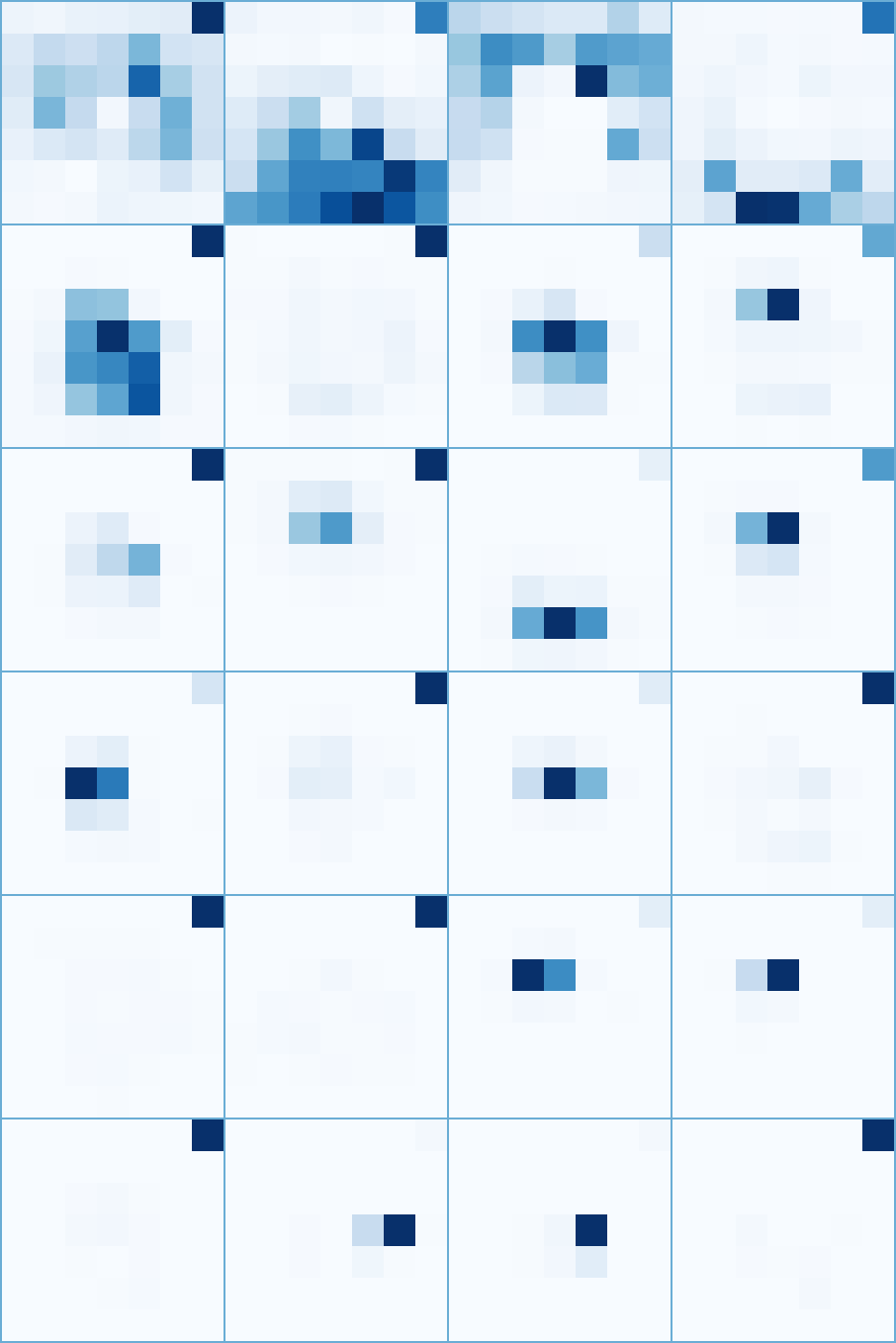} 
		\\			
		(a) DINO: DeiT-S  & 
		\hspace{-2mm}
		(b) \shortname{}: $\Lcal_V$    \vspace{2mm} & 
		\hspace{-2mm}		
		(c) \shortname{}: $\Lcal_V\!+\!\Lcal_R$   \hspace{-0mm} \\ 
	\end{tabular}
	\vspace{-3mm}
	\caption{The learned attention maps for all heads at the top layer, ranked by the entropy of softmax probability. Query is the blue dot in the top-left of the image. Top: Entropy of each heads. Middle: top 60\% probability mass. Bottom: full attention maps.  $\Lcal_R$ shows more attention patterns than $\Lcal_V$ only.
	 }
	\vspace{-1mm}
	\label{fig:attn_appendix_dog}
\end{figure*}

\begin{figure*}[t!]
	\vspace{-0mm}\centering
	\begin{tabular}{c c c}
	\multicolumn{3}{c}{\includegraphics[height=2.6cm]{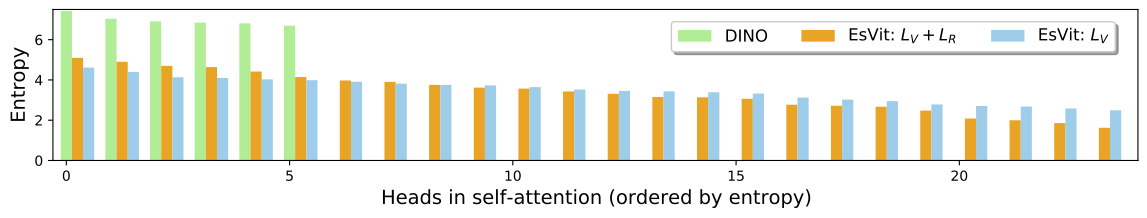}} 	\hspace{-4mm} \\
		\hspace{-2mm}
		\includegraphics[height=7.2cm]{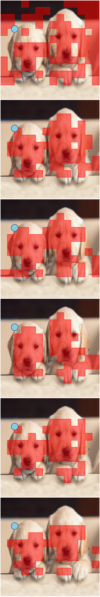}  & 
		\hspace{-3mm}
		\includegraphics[height=7.2cm]{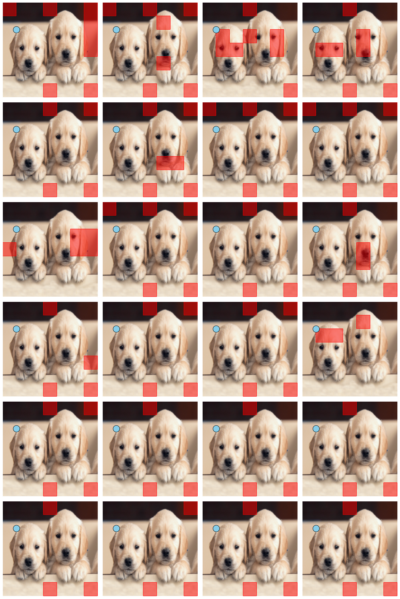} & 
		\hspace{3mm}
		\includegraphics[height=7.2cm]{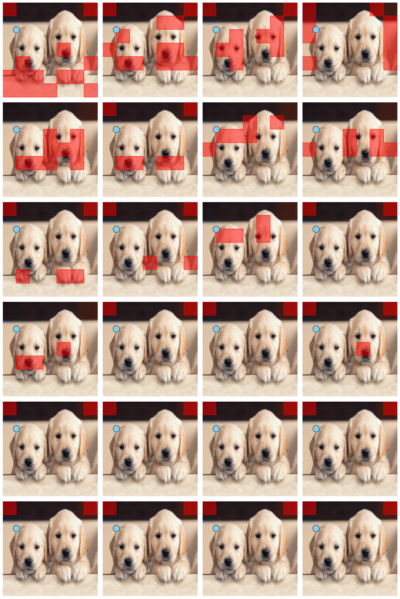} 		
		\\	
		\hspace{-2mm}
		\includegraphics[height=7.2cm]{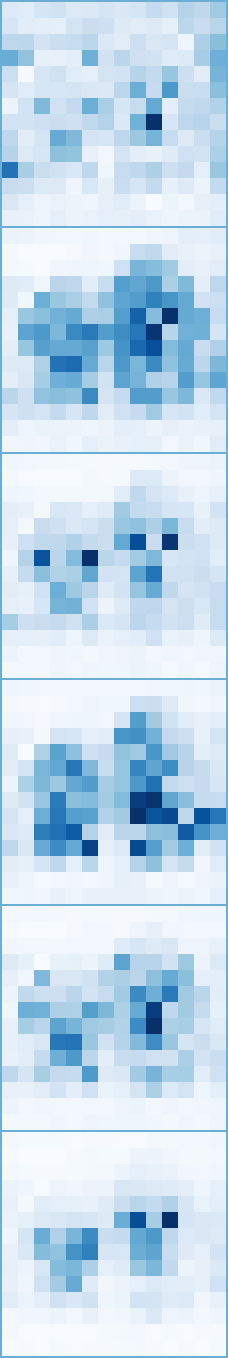}  & 
		\hspace{-2mm}
		\includegraphics[height=7.2cm]{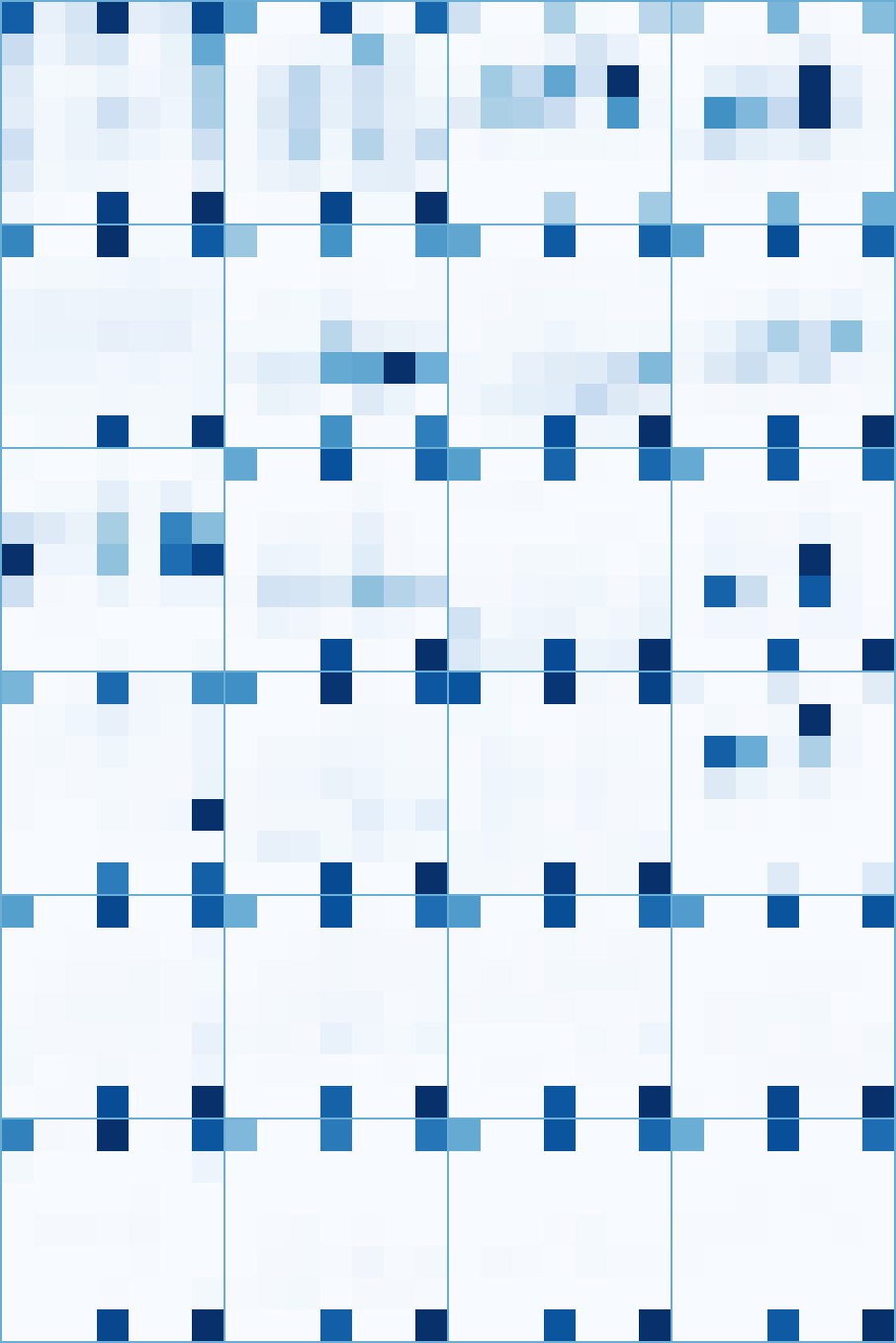} & 
		\hspace{3mm}
		\includegraphics[height=7.2cm]{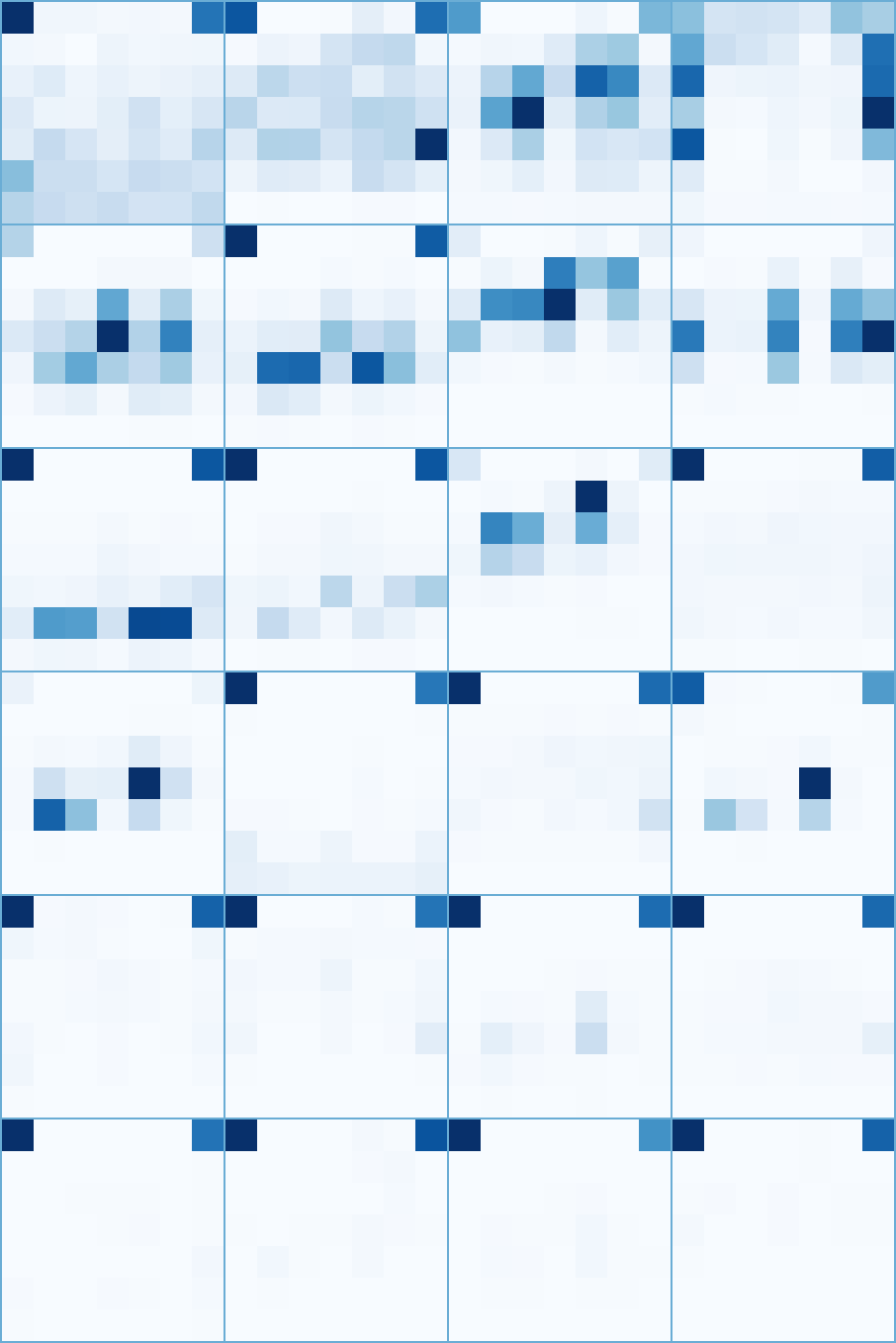} 
		\\			
		(a) DINO: DeiT-S  & 
		\hspace{-2mm}
		(b) \shortname{}: $\Lcal_V$    \vspace{2mm} & 
		\hspace{-2mm}		
		(c) \shortname{}: $\Lcal_V\!+\!\Lcal_R$   \hspace{-0mm} \\ 
	\end{tabular}
	\vspace{-3mm}
	\caption{The learned attention maps for all heads at the top layer, ranked by the entropy of softmax probability. Query is the blue dot in the center of the image. Top: Entropy of each heads. Middle: top 60\% probability mass. Bottom: full attention maps. $\Lcal_R$ shows more attention patterns than $\Lcal_V$ only.
	 }
	\vspace{-1mm}
	\label{fig:attn_appendix_2dogs}
\end{figure*}

\begin{figure*}[t!]
	\vspace{-0mm}\centering
	\begin{tabular}{c c c}
	\multicolumn{3}{c}{\includegraphics[height=2.6cm]{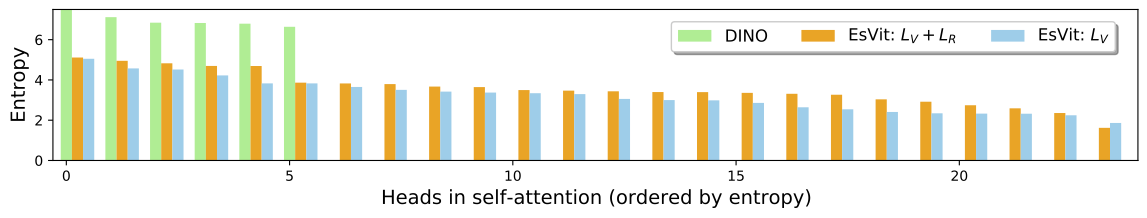}} 	\hspace{-4mm} \\
		\hspace{-2mm}
		\includegraphics[height=7.2cm]{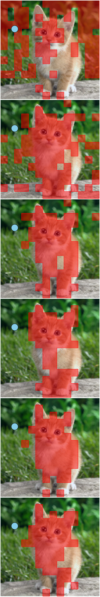}  & 
		\hspace{-3mm}
		\includegraphics[height=7.2cm]{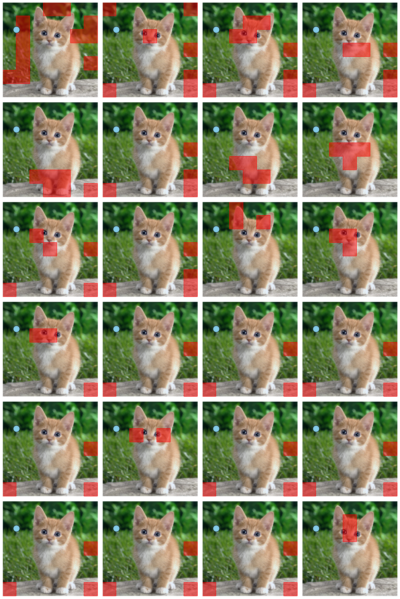} & 
		\hspace{3mm}
		\includegraphics[height=7.2cm]{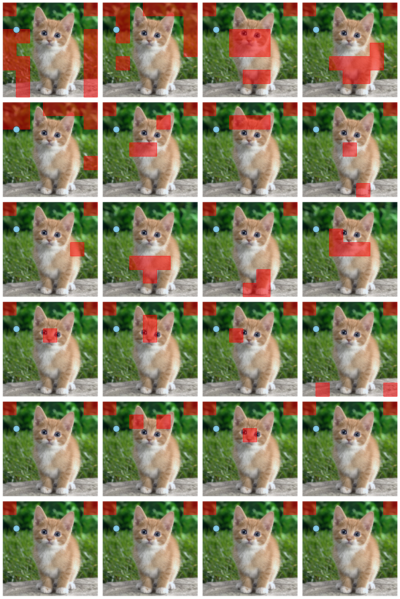} 		
		\\	
		\hspace{-2mm}
		\includegraphics[height=7.2cm]{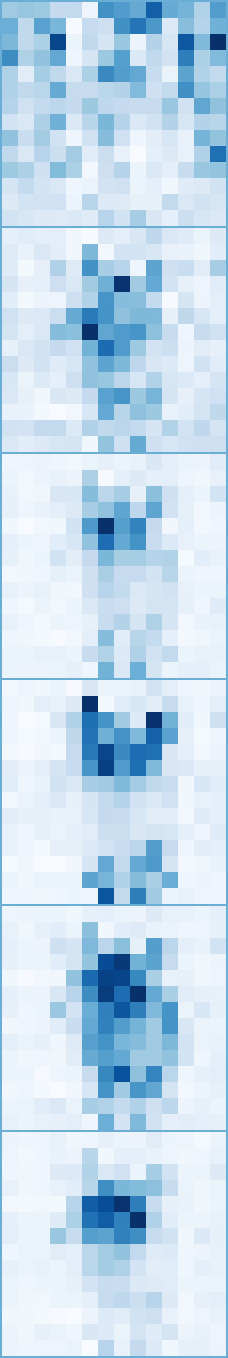}  & 
		\hspace{-2mm}
		\includegraphics[height=7.2cm]{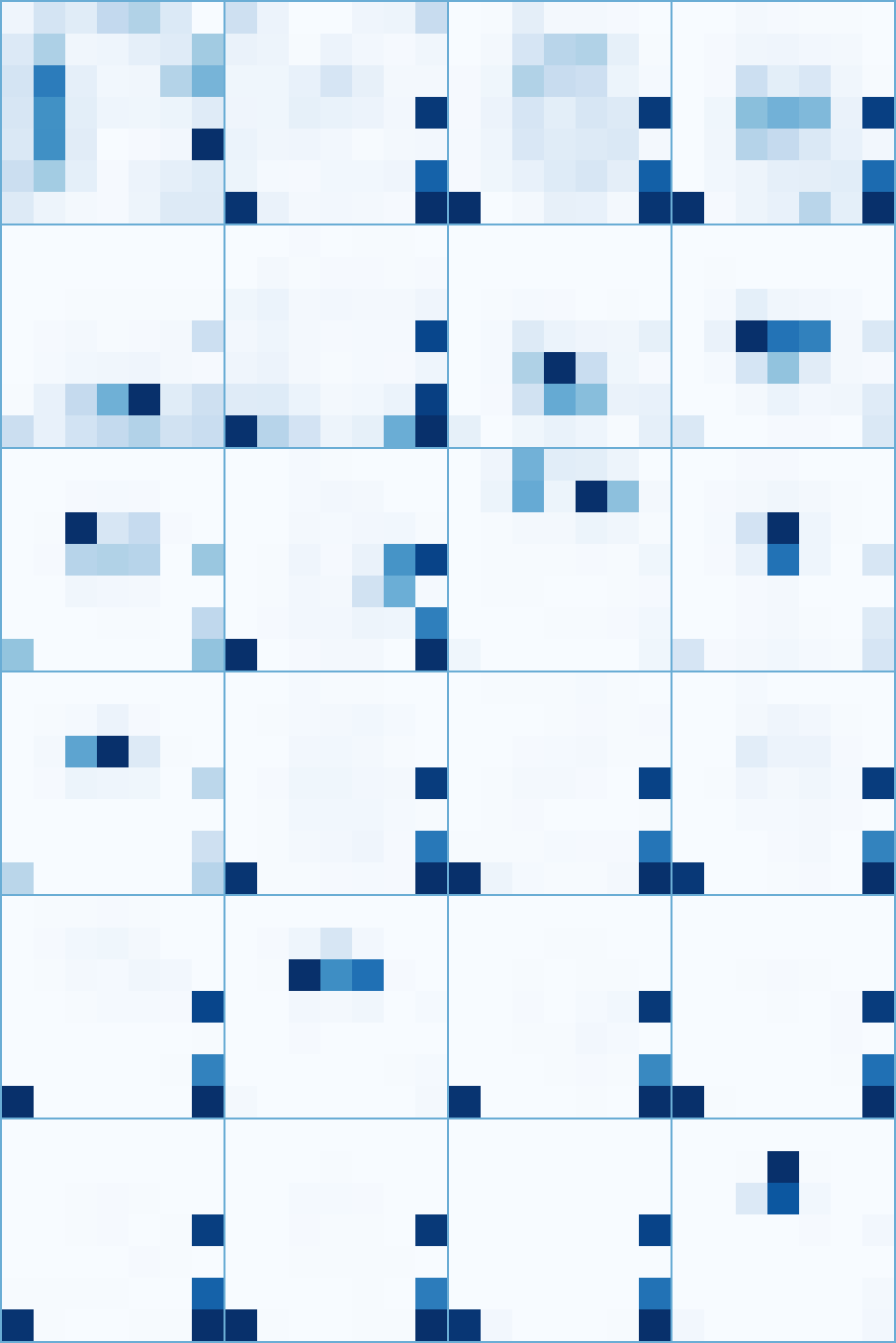} & 
		\hspace{3mm}
		\includegraphics[height=7.2cm]{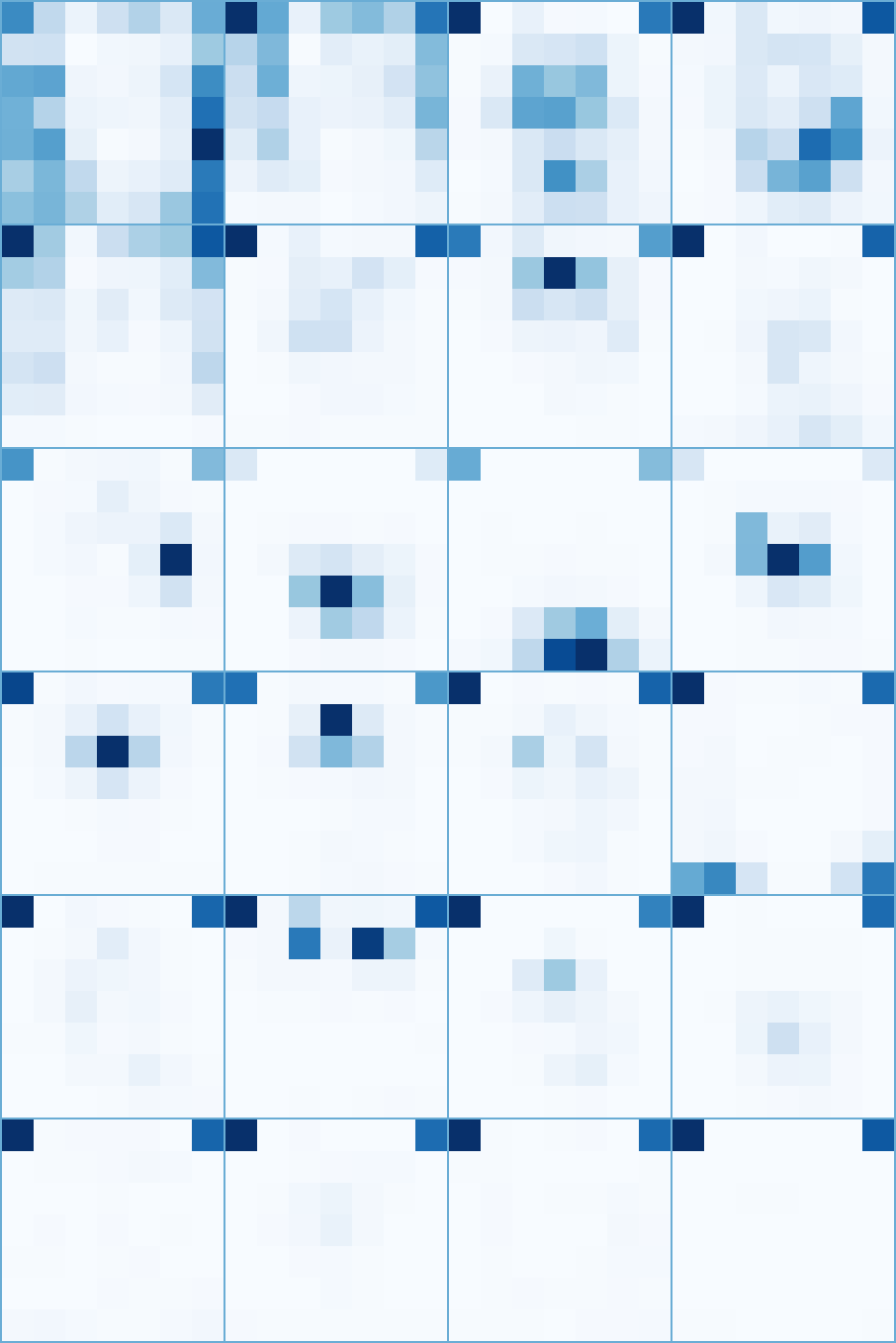} 
		\\			
		(a) DINO: DeiT-S  & 
		\hspace{-2mm}
		(b) \shortname{}: $\Lcal_V$    \vspace{2mm} & 
		\hspace{-2mm}		
		(c) \shortname{}: $\Lcal_V\!+\!\Lcal_R$   \hspace{-0mm} \\ 
	\end{tabular}
	\vspace{-3mm}
	\caption{The learned attention maps for all heads at the top layer, ranked by the entropy of softmax probability. Query is the blue dot in the top-left of the image. Top: Entropy of each heads. Middle: top 60\% probability mass. Bottom: full attention maps. DINO mainly attends the main object even when the query is a background region.
	 }
	\vspace{-1mm}
	\label{fig:attn_appendix_cat}
\end{figure*}

We visualize attention maps at the top layer in Figure~\ref{fig:attn_appendix_dog},~\ref{fig:attn_appendix_2dogs},~\ref{fig:attn_appendix_cat}. 
%
%
With a monolithic Transformer architecture, DINO can automatically identify the main foreground objects. Unfortunately, changing from monolithic to the multi-stage Transformer architecture (From left column to middle column), this property gets lost. There are more heads in the multi-stage architecture than monolithic architecture (24 heads vs 6 heads in this case) in the last year. A fair number of heads in \shortname{} shows redundant patterns, this issue can be reduced when the region-level matching task is added (From middle column to right column).

We observed that DINO with monolithic Transformer architecture only learns to attend the fore-ground objects, even when the query is a background region (see Figure~\ref{fig:attn_appendix_cat}). This is perhaps because DINO models are trained to learn view-level invariance, the main objects in the pre-train dataset ImageNet tend to be the principle factor that remains invariant across different augmented views. Hence, all backgrounds are ignored, regardless of the query positions. This is improved in \shortname{} with the region-level pre-train task, as the model is trained to match individual regions.

DINO shows high entropy values in all of 6 heads (perhaps a required condition to cover all regions of the main object). In \shortname{}, $\Lcal_R$ plays an interesting role in modulating the entropy distributions among heads: it increases those with larger entropy values, while decreasing those with lower entropy values. In another word, it makes the attention patterns in different heads more diverse.

\end{document}